\documentclass{article}

\usepackage{booktabs} 
\usepackage{siunitx} 
\usepackage{subfig}
\usepackage{graphicx}
\usepackage{multirow}
\usepackage{booktabs,colortbl}
\usepackage{soul}
\usepackage{ragged}
\usepackage{pgfplotstable} 
\usepackage{authblk}
\usepackage{geometry}
\usepackage{listings}
\usepackage{lettrine}
\newcommand{\dropcap}[1]{\lettrine[lines=2,lraise=0.05,findent=0.1em, nindent=0em]{{\dropcapfont{#1}}}{}}
 
\makeatletter         
\renewcommand\AB@affilsepx{, \protect\Affilfont}
\renewcommand\maketitle{
{\raggedright 
{\raggedright\baselineskip= 24pt\titlefont \@title\par} \vskip10pt
{\raggedright \@author\par} \vskip8pt
}} 
\makeatother

\renewcommand\Affilfont{\normalfont\sffamily\fontsize{7}{8}\selectfont}

\newcommand{\titlefont}{\fontfamily{lmss}\bfseries\fontsize{22pt}{24pt}\selectfont}
\newcommand{\dropcapfont}{\fontfamily{lmss}\bfseries\fontsize{26pt}{28pt}\selectfont}

\author[a, b, 1]{Ant\^{o}nio H. Ribeiro}
\author[a]{Manoel Horta Ribeiro} 
\author[a, c]{Gabriela M.M. Paix\~{a}o}
\author[a]{Derick M. Oliveira}
\author[a, c]{Paulo R. Gomes}
\author[a, c]{J\'{e}ssica A. Canazart}
\author[a, c]{Milton P. S. Ferreira}
\author[b]{Carl R. Andersson}
\author[d]{Peter W. Macfarlane}
\author[a]{Wagner Meira Jr.}
\author[b, 2]{Thomas B. Sch\"{o}n}
\author[a, c, 3]{Antonio Luiz P. Ribeiro}

\affil[a]{Universidade Federal de Minas Gerais, Brazil}
\affil[b]{Uppsala University, Sweden}
\affil[c]{Telehealth Center from Hospital das Cl\'{i}nicas da Universidade Federal de Minas Gerais, Brazil}
\affil[d]{Glasgow University, Scotland}
\affil[1]{antonio-ribeiro@ufmg.br}
\affil[2]{thomas.schon@it.uu.se}
\affil[3]{tom@hc.ufmg.br}

\title{\textbf{Automatic diagnosis of the 12-lead ECG using a deep neural network}}

\begin{document}

\vspace*{150pt}
\maketitle

\begin{abstract}
The role of automatic electrocardiogram (ECG) analysis in clinical practice is limited by the accuracy of existing models. Deep Neural Networks (DNNs) are models composed of stacked transformations that learn tasks by examples. This technology has recently achieved striking success in a variety of task and there are great expectations on how it might improve clinical practice. Here we present a DNN model trained in a dataset with more than 2 million labeled exams analyzed by the Telehealth Network of Minas Gerais and collected under the scope of the CODE (Clinical Outcomes in Digital Electrocardiology) study. The DNN outperform cardiology resident medical doctors in recognizing 6 types of abnormalities in 12-lead ECG recordings, with F1 scores above $80\%$ and specificity over $99\%$. These results indicate ECG analysis based on DNNs, previously studied in a single-lead setup, generalizes well to 12-lead exams, taking the technology closer to the standard clinical practice.
\end{abstract}

\begin{figure}[b]
\noindent\rule[0.5ex]{\linewidth}{1pt}
{\footnotesize
Preprint. The final version of this paper was published in Nature Communications -- volume: \textbf{11}, article number: 1760 (2020). https://doi.org/10.1038/s41467-020-15432-4.

\begin{lstlisting}[frame=single]
@article{ribeiro_automatic_2020,
  title = {Automatic diagnosis of the 12-lead {{ECG}} using a deep neural network},
  author = {Ribeiro, Ant{\^o}nio H. and Ribeiro, Manoel Horta and Paix{\~a}o, Gabriela M. M. and Oliveira, Derick M. and Gomes, Paulo R. and Canazart, J{\'e}ssica A. and Ferreira, Milton P. S. and Andersson, Carl R. and Macfarlane, Peter W. and Meira Jr., Wagner and Sch{\"o}n, Thomas B. and Ribeiro, Antonio Luiz P.},
  journal = {Nature Communications}
  year = {2020},
  volume = {11},
  number = {1},
  pages = {1760},
  issn = {2041-1723},
  doi = {10.1038/s41467-020-15432-4},
  url = {https://doi.org/10.1038/s41467-020-15432-4},
}
\end{lstlisting}
}

\end{figure}

\newpage

\dropcap{C}ardiovascular diseases are the leading cause of death worldwide~\cite{roth_global_2018} and the electrocardiogram (ECG) is a major tool in their diagnoses. As ECGs transitioned from analog to digital, automated computer analysis of standard 12-lead electrocardiograms gained importance in the process of medical diagnosis~\cite{willems_testing_1987, schlapfer_computerinterpreted_2017}. However, limited performance of classical algorithms~\cite{willems_diagnostic_1991, shah_errors_2007} precludes its usage as a standalone diagnostic tool and relegates them to an ancillary role~\cite{estes_computerized_2013, schlapfer_computerinterpreted_2017}.

Deep neural networks (DNNs) have recently achieved striking success in tasks such as image classification~\cite{krizhevsky_imagenet_2012} and speech recognition~\cite{hinton_deep_2012}, and there are great expectations when it comes to how this technology may improve health care and clinical practice~\cite{stead_clinical_2018, naylorc_prospects_2018, hinton_deep_2018}. So far, the most successful applications used a supervised learning setup to automate diagnosis from exams. Supervised learning models, which learn to map an input to an output based on example input-output pairs, have achieved better performance than a human specialist on their routine work-flow in diagnosing breast cancer~\cite{bejnordi_diagnostic_2017} and detecting retinal diseases from three-dimensional optical coherence tomography scans~\cite{defauw_clinically_2018}.  While efficient, training DNNs in this setup introduces the need for large quantities of labeled data which, for medical applications, introduce several challenges, including those related to confidentiality and security of personal health information~\cite{beck_protecting_2016}.

A convincing preliminary study of the use of DNNs in ECG analysis was recently presented in~\cite{hannun_cardiologistlevel_2019}. For single-lead ECGs, DNNs could match state-of-the-art algorithms when trained in openly available datasets (e.g. 2017 PhysioNet Challenge data~\cite{clifford_af_2017}) and, for a large enough training dataset, present superior performance when compared to practicing cardiologists. However, as pointed out by the authors, it is still an open question if the application of this technology would be useful in a realistic clinical setting, where 12-lead ECGs are the standard technique~\cite{hannun_cardiologistlevel_2019}. 

The short-duration, standard, 12-lead ECG (S12L-ECG) is the most commonly used complementary exam for the evaluation of the heart, being employed across all clinical settings, from the primary care centers to the intensive care units. While long-term cardiac monitoring, such as in the Holter exam, provides information mostly about cardiac rhythm and repolarization, the S12L-ECG can provide a full evaluation of the cardiac electrical activity. This includes arrhythmias, conduction disturbances, acute coronary syndromes, cardiac chamber hypertrophy and enlargement and even the effects of drugs and electrolyte disturbances. Thus, a deep learning approach that allows for accurate interpretation of S12L-ECGs would have the greatest impact.

S12L-ECGs are often performed in settings, such as in primary care centers and emergency units, where there are no specialists to analyze and interpret the ECG tracings. Primary care and emergency department health professionals have limited diagnostic abilities in interpreting  S12-ECGs~\cite{mant_accuracy_2007, veronese_emergency_2016}. The need for an accurate automatic interpretation is most acute in low and middle-income countries, which are responsible for more than $75\%$ of deaths related to cardiovascular disease~\cite{worldhealthorganization_global_2014}, and where the population, often, do not have access to cardiologists with full expertise in ECG diagnosis. 

The use of DNNs for S12L-ECG is still largely unexplored. A contributing factor for this is the shortage of full digital S12L-ECG databases, since most recordings are still registered only on paper, archived as images, or stored in PDF format \cite{sassi_pdfecg_2017}. Most available databases comprise a few hundreds of tracings and no systematic annotation of the full list of ECG diagnoses~\cite{lyon_computational_2018}, limiting their usefulness as training datasets in a supervised learning setting. This lack of systematically annotated data is unfortunate, as training an accurate automatic method of diagnosis from S12L-ECG would be greatly beneficial.

In this paper, we demonstrate the effectiveness of DNNs for automatic S12L-ECG classification. We build a large-scale dataset of labelled S12L-ECG exams for clinical and prognostic studies (the CODE - Clinical Outcomes in Digital Electrocardiology study) and use it to develop a DNN to classify 6 types of ECG abnormalities considered representative of both rhythmic and morphologic ECG abnormalities.

\section{Results}

\subsection{Model specification and training}

We collected a dataset consisting of 2,322,513 ECG records from 1,676,384 different patients of 811 counties in the state of Minas Gerais/Brazil from the  Telehealth Network of Minas Gerais (TNMG)~\cite{alkmim_improving_2012}. The dataset characteristics are summarized in Table~\ref{tab:diagnosis}. The acquisition and annotation procedures of this dataset are described in Methods. We split this dataset into a training set and a validation set. The training set contains 98\% of the data. The validation set consists of the remaining 2\% (\textasciitilde 50,000 exams) of the dataset and it was used for hyperparameter tuning. 

\begin{table}[h]
  \centering
  \begin{tabular}{ccc}
    \hline
    \rowcolor[gray]{0.9} Abnormality & Train+Val (n = 2,322,513) & Test (n = 827)\\
    \hline
    1dAVb  &   35,759  (1.5 \%) &  28 (3.4 \% )\\
    RBBB &  63,528 (2.7\% ) &  34 (4.1 \%)\\
    LBBB & 39,842  (1.7\%) &  30 (3.6 \%)\\
    SB & 37,949 (1.6\%) & 16 (1.9 \%)\\
    AF &  41,862 (1.8\%) &  13 (1.6 \%)\\
    ST & 49,872 (2.1\%) &  36 (4.4 \%) \\
    \hline
    \rowcolor[gray]{0.9} Age group & & \\
    16-25 & 155,531  (6.7 \%) &  43 (5.2 \% )\\
    26-40 & 406,239  (17.5 \%)  &  122 (14.8 \% )\\
    41-60 & 901.456  (38.8 \%) &  340 (41.1 \% )\\
    61-80 & 729,300 (31.4 \%) &  278 (33.6 \% )\\
    $\ge$81 &  129,987 (5.6 \%) &  44 (5.3 \% )\\
    \hline
    \rowcolor[gray]{0.9} Sex & & \\
    Male & 922,780  (39.7 \%) &  321 (38.8 \% )\\
    Female &  1,399,733  (60.3 \%) &  506 (61.2 \% ) \\
    \hline
    \hline
    \end{tabular}
  \caption{\textbf{(Dataset summary)} Patient characteristcs and abnormalities prevalence, n (\%).}
  \label{tab:diagnosis}
\end{table}

\begin{figure}[htpb]
    \centering
    \includegraphics[width=0.6\textwidth]{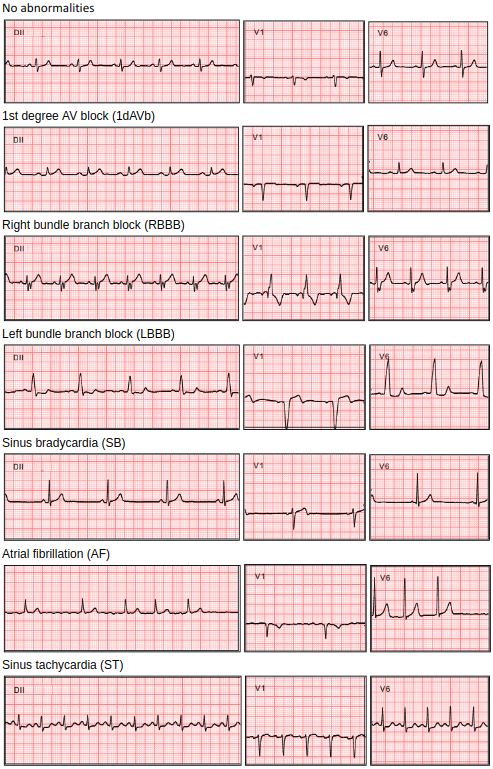}
    \caption{\textbf{(Abnormalities examples)} A list of all the abnormalities the model classifies. We show only 3 representative leads (DII, V1 and V6).}
    \label{fig:abnormalities}
\end{figure}

We train a DNN to detect: 1st degree AV block (1dAVb), right bundle branch block (RBBB), left bundle branch block (LBBB), sinus bradycardia (SB), atrial fibrillation (AF) and sinus tachycardia (ST). These 6 abnormalities are displayed in Figure~\ref{fig:abnormalities}.

We used a DNN architecture known as the residual network~\cite{he_deep_2016}, commonly used for images, which we here have adapted to unidimensional signals. A similar architecture has been successfully employed for detecting abnormalities in single-lead ECG signals~\cite{hannun_cardiologistlevel_2019}. Furthermore, in the  2017 Physionet challenge~\cite{clifford_af_2017}, algorithms for detecting AF have been compared in an open dataset of single lead ECGs and, both the architecture described in~\cite{hannun_cardiologistlevel_2019} and other convolutional architectures~\cite{hong_encase_2017, kamaleswaran_robust_2018} have achieved top scores.

The DNN parameters were learned using the training dataset and our design choices were made in order to maximize the performance on the validation dataset. We should highlight that, despite using a significantly larger training dataset, we got the best validation results with an architecture with, roughly, one quarter the number of layers and parameters of the network employed in~\cite{hannun_cardiologistlevel_2019}.

\subsection{Testing and perfomance evaluation}
\begin{figure*}[h]
    \centering
    \subfloat[][1dAVb]{\includegraphics[width=0.333\textwidth]{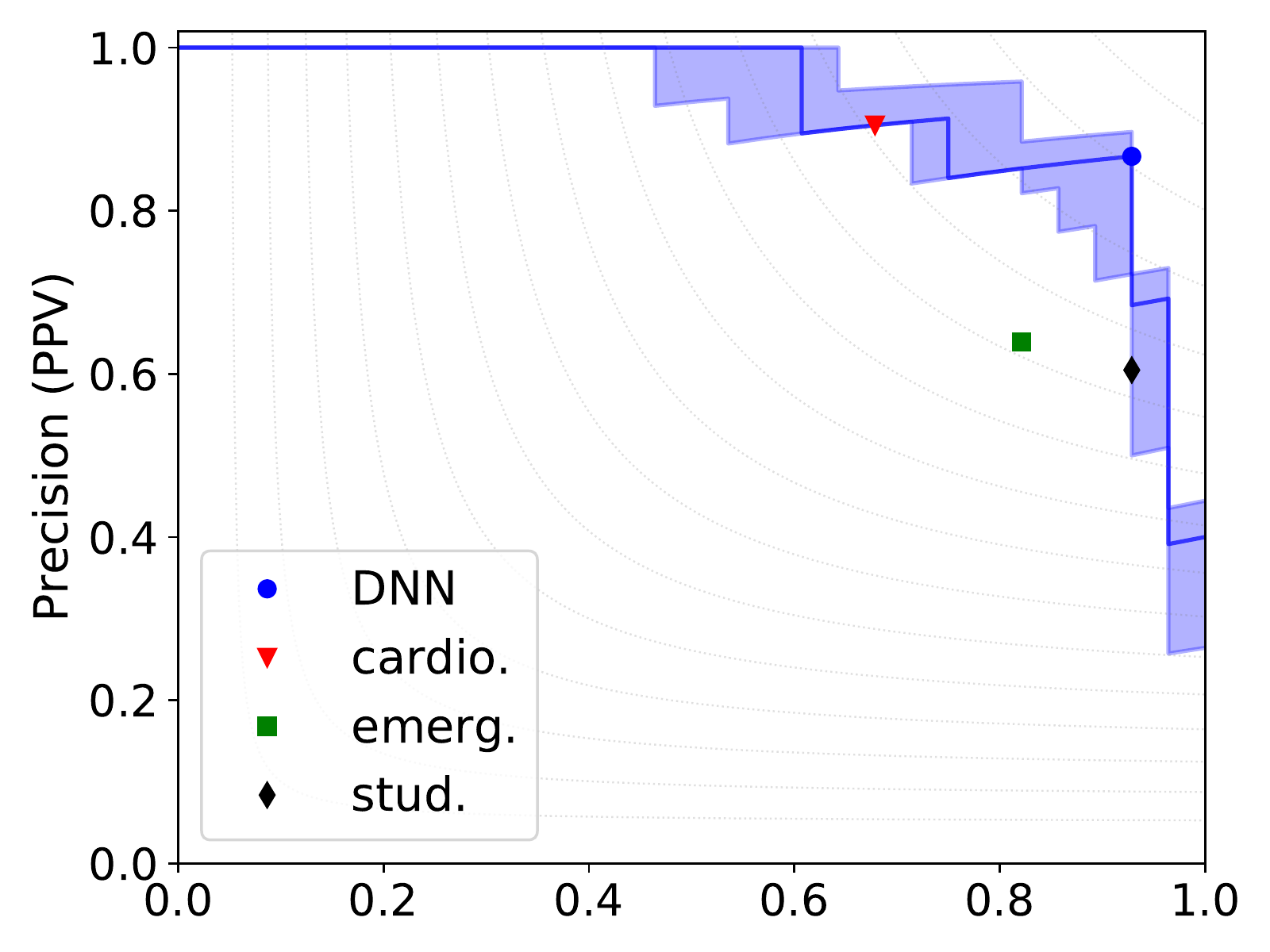}}
    \subfloat[][RBBB]{\includegraphics[width=0.333\textwidth]{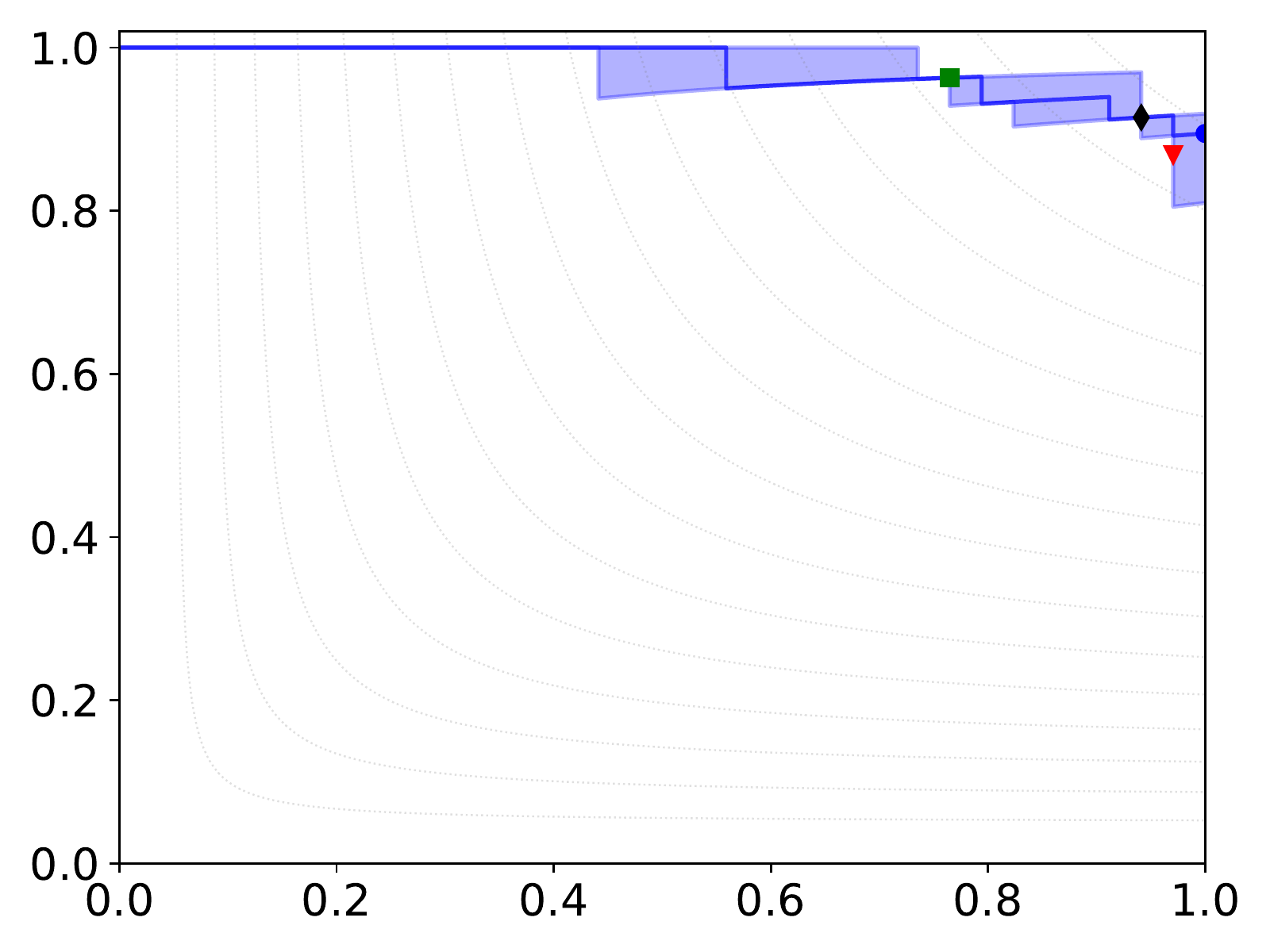}}
    \subfloat[][LBBB]{\includegraphics[width=0.333\textwidth]{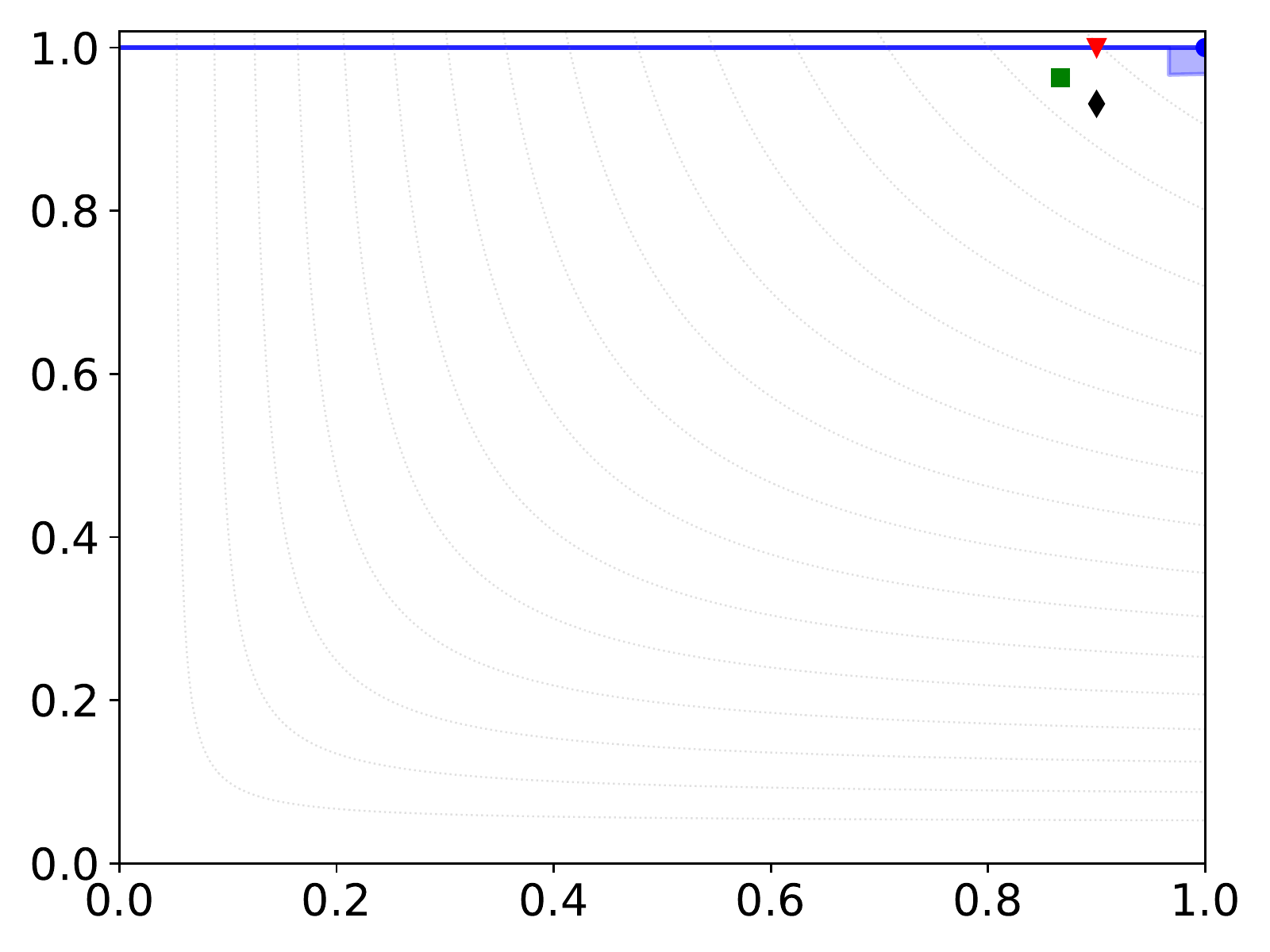}}\\
    \subfloat[][SB]{\includegraphics[width=0.333\textwidth]{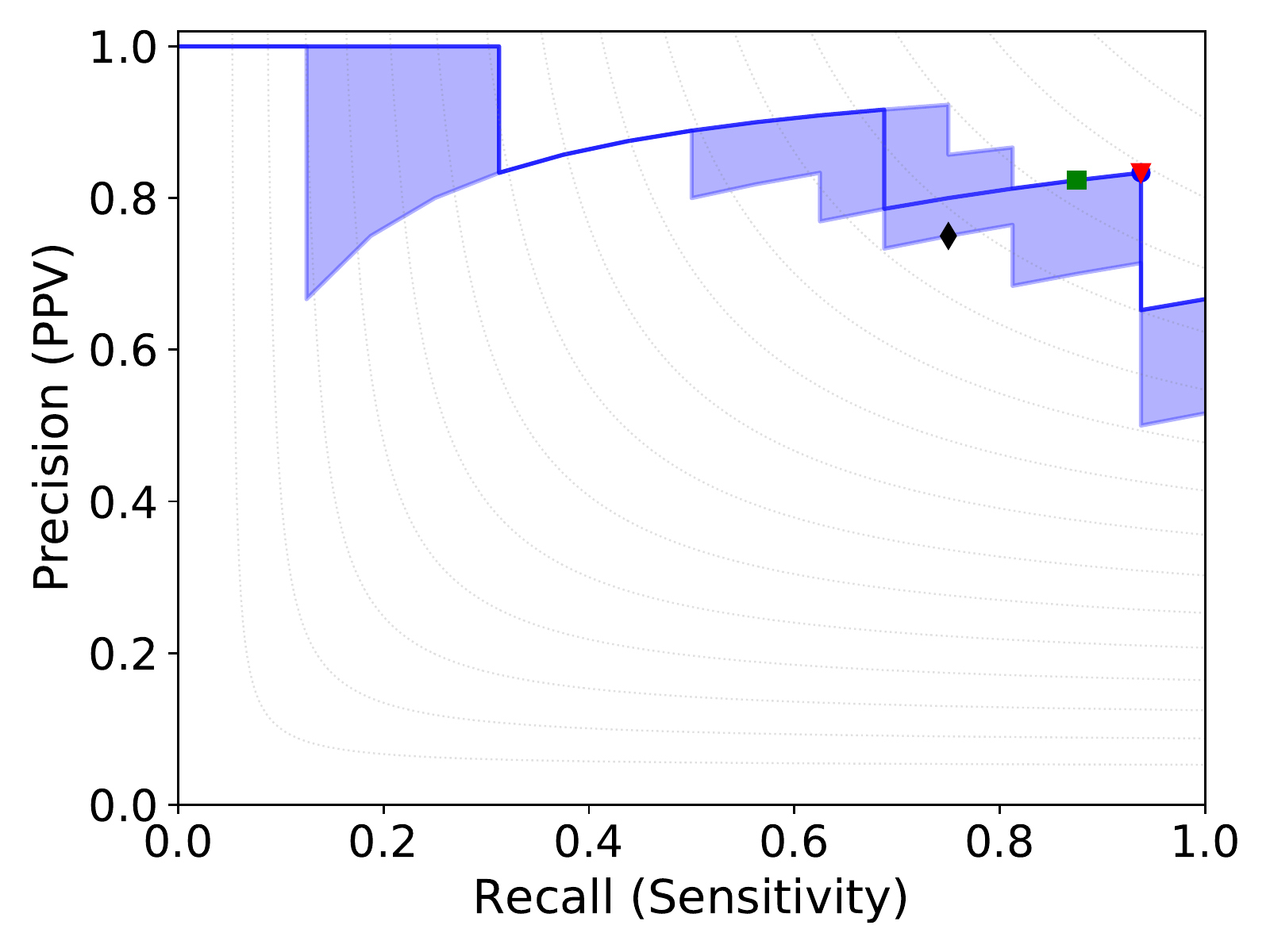}}
    \subfloat[][AF]{\includegraphics[width=0.333\textwidth]{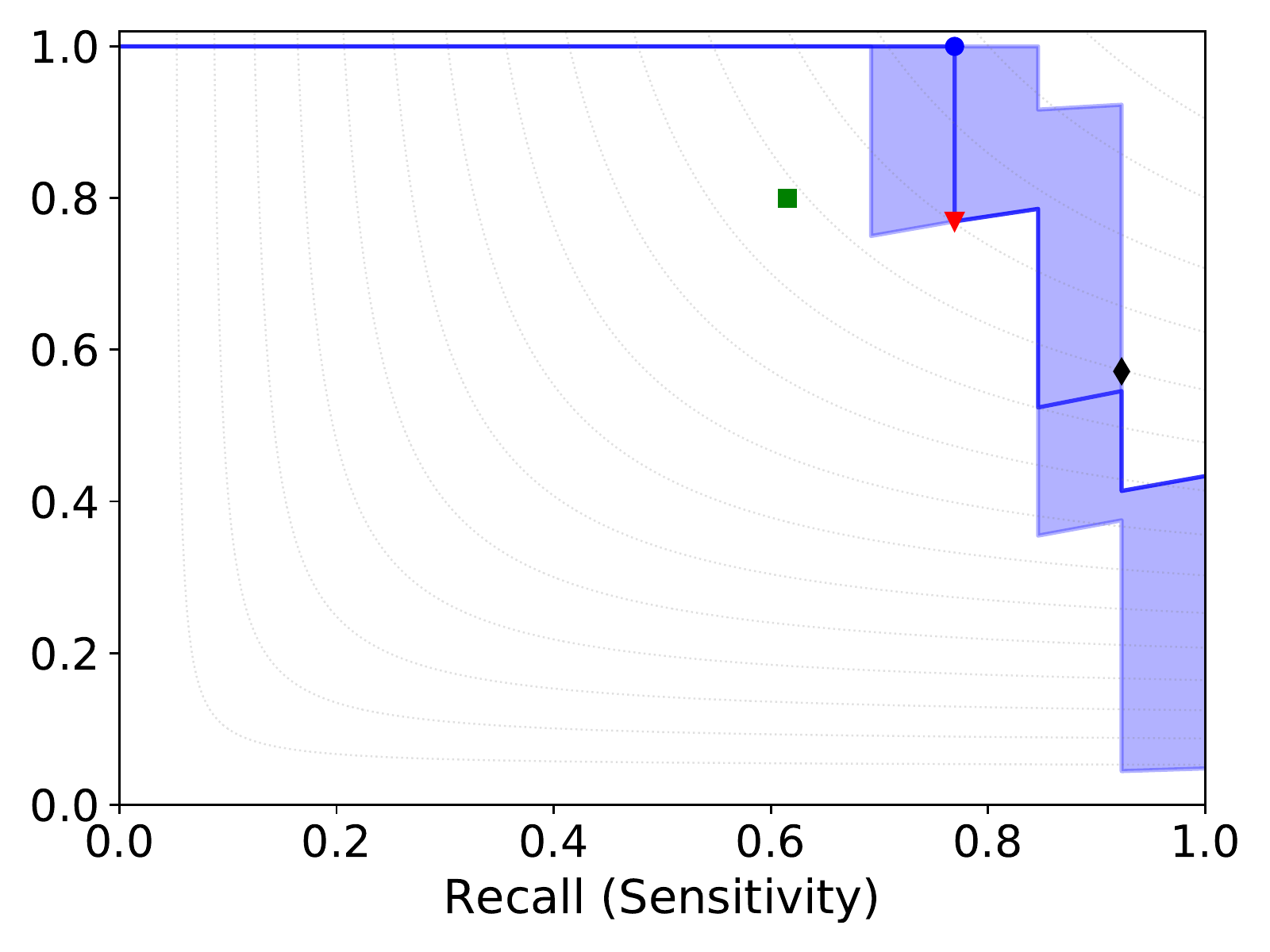}}
    \subfloat[][ST]{\includegraphics[width=0.333\textwidth]{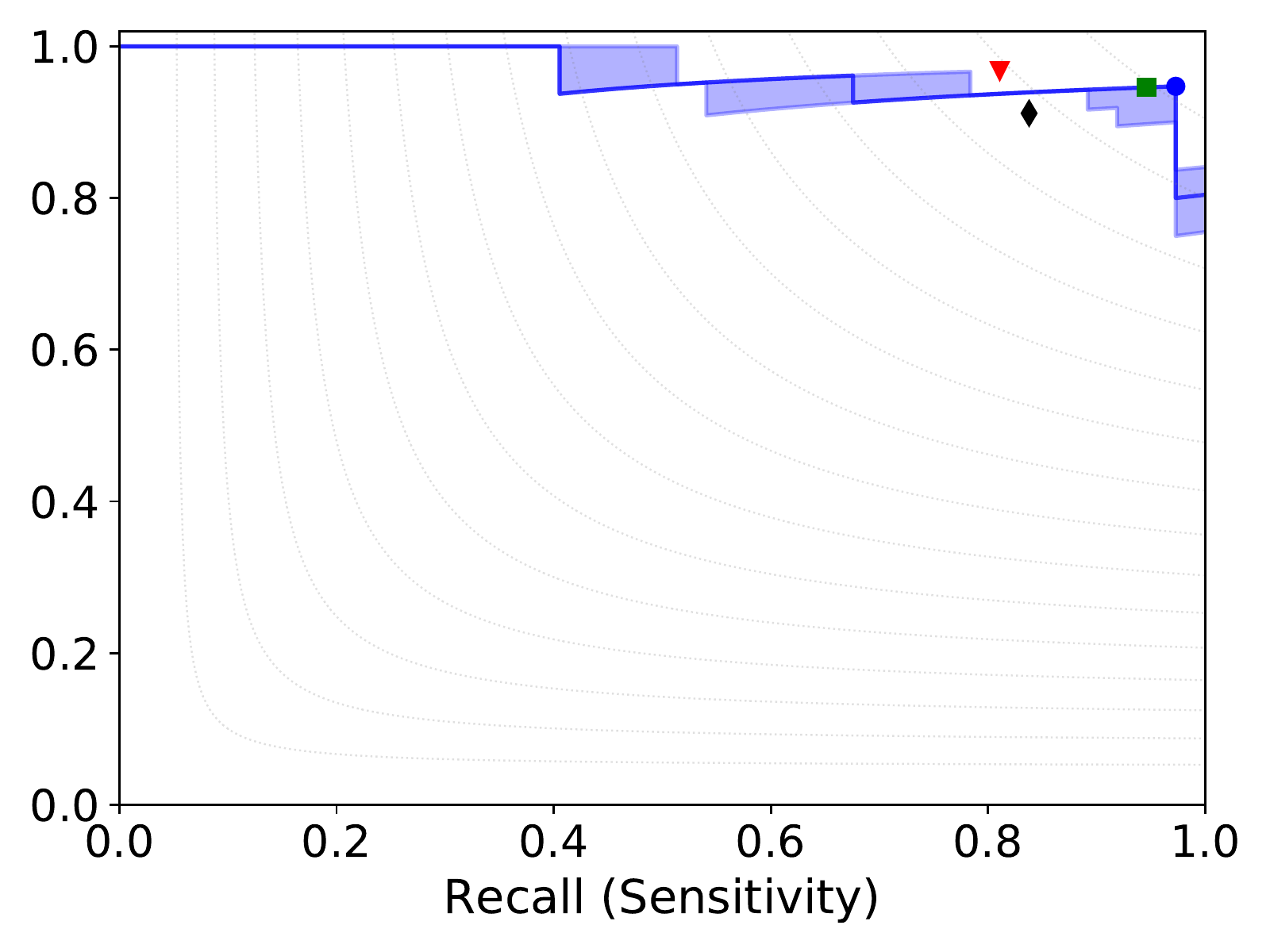}}
    \caption{\textbf{(Precision-recall curve)} Show precision-recall curve for our nominal prediction model on the test set (strong line) with regard to each ECG abnormalities. The shaded region show the range between maximum and minimum precision for neural networks trained with the same configuration and different initialization. Points corresponding the performance of resident medical doctors and students are also displayed, together with the point corresponding to the DNN performance for the same threshold used for generating Table~\ref{tab:performance}. Gray dashed curves in the background correspond to iso-$F_1$ curves (i.e. curves in the precision-recall plane with constant $F_1$ score).}
    \label{fig:precision_recall}
\end{figure*}

For \textit{testing} the model we employed a dataset consisting of $827$ tracings from distinct patients annotated by 3 different cardiologists with experience in electrocardiography (see Methods). The test dataset characteristics are summarized in Table~\ref{tab:diagnosis}. Table~\ref{tab:performance} shows the performance of the DNN on the test set.  High-performance measures were obtained for all ECG abnormalities, with F1 scores above $80\%$ and specificity indexes over $99\%$.  We consider our model to have predicted the abnormality when its output --- a number between 0 and 1 --- is above a threshold. Figure~\ref{fig:precision_recall} shows the precision-recall curve for our model, for different values of this threshold.

Neural networks are initialized randomly, and different initialization usually yield different results. In order to show the stability of the method, we have trained 10 neural networks with the same set of hyperparameters and different initializations. The range between the maximum and minimum precision among these realizations, for different values of thereshold, are the shaded regions displayed in Figure~\ref{fig:precision_recall}. These realizations have micro average precision (mAP) between 0.946 and 0.961, we choose the one with mAP imediatly above the median value of all executions (the one with mAP = 0.951)\footnote{We couldn't choose the model with mAP equal to the median value because 10 is even number, hence there is no single middle value.}. All the analysis from now on will be for this realization of the neural network, which correspond both to the strong line in Figure~\ref{fig:precision_recall} and to the scores presented in Table~\ref{tab:performance}. For this model, Figure~\ref{fig:precision_recall} shows the point corresponding to the maximum F1 score for each abnormality. The threshold corresponding to this point is used for producing the DNN scores displayed in Table~\ref{tab:performance}.

The same dataset was evaluated by: i) two 4th year cardiology residents; ii) two 3rd year emergency residents; and, iii) two 5th year medical students. Each one annotated half of the exams in the test set. Their average performances are given, together with the DNN results, in the Table~\ref{tab:performance} and their precision-recall scores are plotted on Figure~\ref{fig:precision_recall}.  Considering the F1 score, the DNN matches or outperforms the medical residents and students for all abnormalities. The confusion matrices and the inter-rater agreement (kappa coefficients) for the DNN, the resident medical doctors and students are provided, respectively, in  Supplementary Tables 1 and~2(a). Additionally, in Supplementary Table~2(b) we compare the inter-rater agreement between the neural network and the certified cardiologists that annotated the test set.

\begin{table*}[t]
    \setlength{\tabcolsep}{4pt}
    \renewcommand{\arraystretch}{1.0}
    \scriptsize
    \centering
\makebox[\textwidth][c]{
        \begin {tabular}{c|cccc|cccc|cccc|cccc}%
    \toprule \multicolumn {1}{c}{}& \multicolumn {4}{c}{Precision (PPV)} & \multicolumn {4}{c}{Recall (Sensitivity)} & \multicolumn {4}{c}{Specificity} & \multicolumn {4}{c}{F1 Score}\\&DNN&cardio.&emerg.&stud.&DNN&cardio.&emerg.&stud.&DNN&cardio.&emerg.&stud.&DNN&cardio.&emerg.&stud.\\\midrule %
    1dAVb& {0.867}& {0.905}& {0.639}& {0.605}& {0.929}& {0.679}& {0.821}& {0.929}& {0.995}& {0.997}& {0.984}& {0.979}&\textbf {0.897}& {0.776}& {0.719}& {0.732}\\%
    RBBB& {0.895}& {0.868}& {0.963}& {0.914}& {1.000}& {0.971}& {0.765}& {0.941}& {0.995}& {0.994}& {0.999}& {0.996}&\textbf {0.944}& {0.917}& {0.852}& {0.928}\\%
    LBBB& {1.000}& {1.000}& {0.963}& {0.931}& {1.000}& {0.900}& {0.867}& {0.900}& {1.000}& {1.000}& {0.999}& {0.997}&\textbf {1.000}& {0.947}& {0.912}& {0.915}\\%
    SB& {0.833}& {0.833}& {0.824}& {0.750}& {0.938}& {0.938}& {0.875}& {0.750}& {0.996}& {0.996}& {0.996}& {0.995}&\textbf {0.882}&\textbf {0.882}& {0.848}& {0.750}\\%
    AF& {1.000}& {0.769}& {0.800}& {0.571}& {0.769}& {0.769}& {0.615}& {0.923}& {1.000}& {0.996}& {0.998}& {0.989}&\textbf {0.870}& {0.769}& {0.696}& {0.706}\\%
    ST& {0.947}& {0.968}& {0.946}& {0.912}& {0.973}& {0.811}& {0.946}& {0.838}& {0.997}& {0.999}& {0.997}& {0.996}&\textbf {0.960}& {0.882}& {0.946}& {0.873}\\\bottomrule %
    \end {tabular}%
        }
    \caption{\textbf{(Performance indexes)} Scores of our DNN are compared on the test set with the average performance of: i) 4th year cardiology resident (\textit{cardio.}); ii) 3rd year emergency resident (\textit{emerg.}); and, iii) 5th year medical students (\textit{stud.}).
    (PPV = positive predictive value)}
    \label{tab:performance}
\end{table*}

A trained cardiologist reviewed all the mistakes made by the DNN, the medical residents and the students, trying to explain the source of the error. The cardiologist had meetings with the residents and students where they together agreed on which was the source of the error. The results of this analysis are given in Table~\ref{tab:error_analysis}.

In order to compare of the performance difference between the DNN and resident medical doctors and students, we compute empirical distributions for the precision (PPV), recall (sensitivity), specificity and F1 score using bootstraping~\cite{efron_introduction_1994}.  The boxplots corresponding to these bootstrapped distributions are presented in Supplementary Figure 1. We have also applied the McNemar test~\cite{mcnemar_note_1947} to compare the misclassification distribution of the DNN, the medical residents and the students. Supplementary Table 3 show the $p$-values of the statistical test. Both analyses do not indicate a statistically significant difference in performance among the DNN and the medical residents and students for most of the classes. 

Finally, to asses the effect of how we structure our problem, we have considered alternative scenarios where we use the 2,322,513 ECG records in 90\%-5\%-5\% splits, stratified randomly, by patient or in chronological order. Being the splits used, respectively, for training, validation and as a second larger \textit{test} set. The results indicate no  statistically significant difference between the original DNN used in our analysis and the alternative models developed in the 90\%-5\%-5\% splits. The exception is the model developed using the chronologically order split, for which the changes along time in the telehealth center operation have affected the splits (cf. Supplementary Figure~2).

\begin{table*}
    \setlength{\tabcolsep}{5pt}
    \renewcommand{\arraystretch}{1.0}
    \scriptsize
    \centering
    \begin {tabular}{c|ccc|cccc|cccc|cccc}%
    \toprule \multicolumn {1}{c}{}& \multicolumn {3}{c}{DNN} & \multicolumn {4}{c}{cardio.} & \multicolumn {4}{c}{emerg.} & \multicolumn {4}{c}{stud.} \\&meas.&noise&unexplain.&meas.&noise&concep.&atte.&meas.&noise&concep.&atte.&meas.&noise&concep.&atte.\\\midrule %
    1dAVb& {3}& {2}& {1}& {8}& {3}&&& {15}& {3}&&& {13}& {3}& {3}&\\%
    RBBB& {3}&& {1}& {4}&& {2}&& {1}&& {8}&& {3}&& {2}&\\%
    LBBB&&&& {1}& {1}& {1}&&& {1}& {4}&&& {2}& {3}&\\%
    SB& {4}&&& {4}&&&& {4}&&& {1}& {5}&& {2}& {1}\\%
    AF&& {2}& {1}&& {4}& {2}&&& {2}& {5}&&& {3}& {7}&\\%
    ST& {2}&& {1}& {2}& {1}&& {5}& {1}& {1}& {1}& {1}& {1}& {2}& {1}& {5}\\\bottomrule %
    \end {tabular}%
    \caption{\textbf{(Error analysis)} Present the analysis of misclassified exams. The errors were classified into the following categories: i) measurements errors (\textit{meas.}) were ECG interval measurements preclude the given diagnosis by its textbook definition ; ii) errors due to \textit{noise}, were we believe that the analyst or the DNN failed due to a lower than usual signal quality; and, iii) other type of errors (\textit{unexplain.}). Those were further divided, for the medical residents and students, into two categories: conceptual errors (\textit{concep.}), where our reviewer suggested that the doctor failed to understand the definitions of each abnormality, and attention errors (\textit{atte.}), where we believe the error could be avoided if the reviewer had been more careful.}
    \label{tab:error_analysis}
\end{table*}

\section{Discussion}

This paper demonstrates the effectiveness of "end-to-end" automatic S12L-ECG classification. This presents a paradigm shift from the classical ECG automatic analysis methods~\cite{jambukia_classification_2015}. These classical methods, such as the University of Glasgow ECG analysis program~\cite{macfarlane_university_2005}, first extract the main features of the ECG signal using traditional signal processing techniques and then use these features as inputs to a classifier. End-to-end learning presents an alternative to these two-step approaches, where the raw signal itself is used as an input to the classifier which learns, by itself, to extract the features. This approach have presented, in a emergency room setting, performance superior to commercial ECG software based on traditional signal processing techniques~\cite{smith_deep_2019}

Neural networks have previously been used for classification of ECGs both in a classical --- feature-based --- setup~\cite{cubanski_neural_1994, tripathy_novel_2019} and in an end-to-end learn setup~\cite{rubin_densely_2017, acharya_application_2017, hannun_cardiologistlevel_2019}. Hybrid methods combining the two paradigms are also available: the classification may be done using a combination of handcrafted and learned features~\cite{shashikumar_detection_2018} or by using a two-stage training, obtaining one neural network to learn the features and another to classify the exam according to these learned features~\cite{rahhal_deep_2016}. 

The paradigm shift towards end-to-end learning had a significant impact on the size of the datasets used for training the models. Many results using classical methods~\cite{jambukia_classification_2015, acharya_application_2017, rahhal_deep_2016} train their models on datasets with few examples, such as the MIT-BIH arrhythmia database~\cite{goldberger_physiobank_2000}, with only 47 unique patients. The  most convincing papers using end-to-end deep learning or mixed approaches, on the other hand, have constructed large datasets, ranging from 3,000 to 100,000 unique patients, for training their models~\cite{shashikumar_detection_2018, clifford_af_2017, hannun_cardiologistlevel_2019, smith_deep_2019}. 

Large datasets from previous work~\cite{shashikumar_detection_2018, clifford_af_2017, hannun_cardiologistlevel_2019}, however either were obtained from cardiac monitors and Holter exams, where patients are usually monitored for several hours and the recordings are restricted to one or two leads. Or, consist of 12-lead ECGs obtained in a emergency room setting~\cite{goto_artificial_2019, smith_deep_2019}. Our dataset with well over 2 million entries, on the other hand, consists of short duration (7 to 10 seconds) S12L-ECG tracings obtained from in-clinic exams and is orders of magnitude larger than those used in previous studies. It encompasses not only rhythm disorders, like AF, SB and ST, as in previous studies~\cite{hannun_cardiologistlevel_2019}, but also conduction disturbances, such as 1dAVb, RBBB and LBBB. Instead of beat to beat classification, as in the MIT-BIH arrhythmia database, our dataset provides annotation for S12L-ECG exams, which are the most common in clinical practice.

The availability of such a large database of S12L-ECG tracings, with annotation for the whole spectrum of ECG abnormalities, opens up the possibility of extending initial results of end-to-end DNN in ECG automatic analysis~\cite{hannun_cardiologistlevel_2019} to a system with applicability in a wide range of clinical settings. The development of such technologies may yield high-accuracy automatic ECG classification systems that could save clinicians considerable time and prevent wrong diagnoses. Millions of S12L-ECGs are performed every year, many times in places where there is a shortage of qualified medical doctors to interpret them. An accurate classification system could help to detect wrong diagnoses and improve the access of patients from deprived and remote locations to this essential diagnostic tool of cardiovascular diseases.

The error analysis shows that most of the DNN mistakes were related to measurements of ECG intervals. Most of those were borderline cases, where the diagnosis relies on a consensus definitions \cite{rautaharju_aha_2009} that can only be ascertained when a measurement is above a sharp cutoff point. The mistakes can be explained by the DNN failing to encode these very sharp thresholds. For example, the DNN wrongly detecting a SB with a heart rate slightly above 50 bpm or a ST with a heart rate slightly below 100 bpm. Supplementary Figure~3 illustrate this effect. Noise and interference in the baseline are established causes of error~\cite{luo_review_2010}  and affected both automatic and manual diagnosis of ECG abnormalities. Nevertheless, the DNN seems to be more robust to noise and it made fewer mistakes of this type compared to the medical residents and students. Conceptual errors (where our reviewer suggested that the doctor failed to understand the definitions of each abnormality) were more frequent for emergency residents and medical students than for cardiology residents. Attention errors (where we believe that the error could have been avoided if the manual reviewer were more careful) were present at a similar ratio for cardiology residents, emergency residents and medical students.

Interestingly, the performance of the emergency residents is worse than medical students for many abnormalities. This might seem counter-intuitive because they do have less years of medical training. It might, however, be justified by the fact that emergency residents, unlike cardiology residents, do not have to interpret these exams on a daily basis, while medical students still have these concepts fresh from their studies.

Our work is perhaps best understood in the context of its limitations. While we obtained the highest F1 scores for the DNN, the McNemar statistical test and bootstrapping suggest that we do not have confidence enough to assert that the DNN is actually better than the medical residents and students with statistical significance. We attribute this lack of confidence in the comparison to the presence of relatively infrequent classes, where a few erroneous classifications may significantly affect the scores. Furthermore, we did not test the accuracy of the DNN in the diagnosis of other classes of abnormalities, like those related to acute coronary syndromes or cardiac chamber enlargements and we cannot extend our results to these untested clinical situations.  Indeed, the real clinical setting is  more complex than the experimental situation tested in this study and, in complex and borderline situations, ECG interpretation can be extremely difficult and may demand the input of highly specialized personnel. Thus, even if a DNN is able to recognize typical ECG abnormalities, further analysis by an experienced specialist will continue to be necessary to these complex exams. 

This proof-of-concept study, showing that a DNN can accurately recognize ECG rhythm and morphological abnormalities in clinical S12L-ECG exams, opens a series of perspectives for future research and clinical applications. A next step would be to prove that a DNN can effectively diagnose multiple and complex ECG abnormalities, including myocardial infarction, cardiac chamber enlargement and hypertrophy and less common forms of arrhythmia, and to recognize a normal ECG. Subsequently, the algorithm should be tested in a controlled real-life situation, showing that accurate diagnosis could be achieved in real-time, to be reviewed by clinical specialists with solid experience in ECG diagnosis. This real-time, continuous evaluation of the algorithm, would provide rapid feedback that could be incorporated as further improvements of the DNN, making it even more reliable.

The TNMG, the large telehealth service from which the dataset used was obtained~\cite{alkmim_improving_2012}, is a natural laboratory for these next steps, since it performs more than 2,000 ECGs a day and it is currently expanding its geographical coverage over a large part of a continental country (Brazil). An optimized system for ECG interpretation, where most of  the classification decisions are made automatically would imply that the cardiologists would only be needed for the more complex cases. If such a system is made widely available, it could be of striking utility to improve access to health care in low and middle-income countries, where cardiovascular diseases are the leading cause of death and systems of care for cardiac diseases are lacking or not working well~\cite{nascimento_implementing_2019}.

In conclusion, we developed an end-to-end DNN capable of accurately recognizing six ECG abnormalities in S12L-ECG exams, with a diagnostic performance at least as good as medical residents and students. This study shows the potential of this technology, which, when fully developed, might lead to more reliable automatic diagnosis and improved clinical practice. Although expert review of complex and borderline cases seems to be necessary even in this future scenario, the development of such automatic interpretation by a DNN algorithm may expand the access of the population to this basic and useful diagnostic exam.

\section{Methods}

\subsection{Dataset acquisition}
All S12L-ECGs analyzed in this study were obtained by the Telehealth Network of Minas Gerais (TNMG), a public telehealth system assisting 811 out of the 853 municipalities in the state of Minas Gerais,  Brazil~\cite{alkmim_improving_2012}. Since September 2017, the TNMG has also provided telediagnostic services to other Brazilian states in the Amazonian and Northeast regions. The S12L-ECG exam was performed mostly in primary care facilities  using a tele-electrocardiograph manufactured by Tecnologia Eletrônica Brasileira (São Paulo, Brazil) – model TEB ECGPC - or Micromed Biotecnologia (Brasilia, Brazil) - model ErgoPC 13. The duration of the ECG recordings is between 7 and 10 seconds sampled at frequencies ranging from 300 to 600 Hz. A specific software developed in-house was used to capture ECG tracings, to upload the exam together with the patient’s clinical history and to send it electronically to the TNMG analysis center. Once there, one cardiologist from the TNMG experienced team analyzes the exam and a report is made available to the health service that requested the exam through an online platform.

We have incorporated the University of Glasgow (Uni-G) ECG analysis program (release 28.5, issued in January 2014) in the in-house software since December 2017. The  analysis program was used to automatically identify waves and to calculate axes, durations, amplitudes and intervals, to perform rhythm analysis and to give diagnostic interpretation~\cite{macfarlane_methodology_1990, macfarlane_university_2005}. The Uni-G analysis program also provides Minnesota codes~\cite{macfarlane_automated_1996}, a standard ECG classification used in epidemiological studies~\cite{prineas_minnesota_2009}. Since April 2018 the automatic measurements are being shown to the cardiologists that give the medical report. All clinical information, digital ECGs tracings and the cardiologist report were stored in a database. All previously stored data was also analyzed by Uni-G software in order to have measurements and automatic diagnosis for all exams available in the database, since the first recordings.  The CODE study was established to standardize and consolidate this database for clinical and epidemiological studies. In the present study, the data (for patients above 16 years old) obtained between 2010 and 2016 was used in the training and validation set and, from April to September 2018, in the test set.

\subsection{Labelling training data from text report}
\label{sec:labelling}

For the training and validation sets, the cardiologist report is available only as a textual description of the abnormalities in the exam.  We extract the label from this textual report using a three-step procedure. First, the text is pre-processed by removing stop-words and generating n-grams from the medical report. Then, the \textit{Lazy Associative Classifier} (LAC)~\cite{veloso_lazy_2006}, trained on a 2800-sample dictionary created from real diagnoses text reports, is applied to the n-grams. Finally, the text label is obtained using the LAC result in a rule-based classifier for class disambiguation. The classification model reported above was tested on 4557 medical reports manually labeled by a certified cardiologist who was presented with the free-text and was required to choose among the pre-specified classes. The classification step recovered the true medical label with good results,  the macro F1 score achieved were: 0.729 for 1dAVb; 0.849 for RBBB; 0.838 for LBBB; 0.991 for SB; 0.993 for AF; 0.974 for ST.

\subsection{Training and validation set annotation}

To annotate the training and validation datasets, we used: i) the Uni-G statements and Minnesota codes obtained by the Uni-G automatic analysis (\textit{automatic diagnosis}); ii) automatic measurements provided by the Uni-G software; and, iii) the text labels extracted from the expert text reports using the semi-supervised methodology (\textit{medical diagnosis}). Both the automatic  and medical diagnosis are subject to errors: automatic classification has limited accuracy~\cite{willems_diagnostic_1991, shah_errors_2007, schlapfer_computerinterpreted_2017, estes_computerized_2013} and text labels are subject both to errors of the practicing expert  cardiologists and the labeling methodology. Hence, we combine the expert annotation with the automatic analysis to improve the quality of the dataset. The following procedure is used for obtaining the ground truth annotation:

\begin{enumerate}
\item We:
\begin{enumerate}
    \item \textit{Accept a diagnosis} (consider an abnormality to be present) if both the expert \emph{and} either the Uni-G statement or the Minnesota code provided by the automatic analysis indicated the same abnormality. 
    \item \textit{Reject a diagnosis} (consider an abnormality to be absent) if only one automatic classifier indicates the abnormality in disagreement with both the doctor and the other automatic classifier. 
\end{enumerate}
After this initial step, there are two scenarios where we still need to accept or reject diagnoses. They are: 
i) both classifiers indicate the abnormality, but the expert does not; or 
ii) only the expert indicates the abnormality, whereas none of the classifiers indicates anything.

\item We used the following rules \textit{to reject some of the remaining diagnoses}:

\begin{enumerate}

    \item Diagnoses of ST where the heart rate was below $100$ ($8376$ medical diagnoses and $2$ automatic diagnoses)  were \textit{rejected}. 
    \item Diagnoses of SB where the heart rate was above $50$ ($7361$ medical diagnoses and $16427$ automatic diagnosis) were \textit{rejected}. 
    \item Diagnoses of LBBB or RBBB where the duration of the QRS interval was below $115$ ms ($9313$ medical diagnoses for RBBB and $8260$ for LBBB)  were \textit{rejected}. 
    \item Diagnoses of 1dAVb where the duration of the PR interval was below $190$ ms ($3987$ automatic diagnoses) were \textit{rejected}.
\end{enumerate}

\item Then, using the sensitivity analysis of $100$ manually reviewed exams per abnormality, we came up with the following rules \textit{to accept some of the remaining diagnoses}:
\begin{enumerate}
    \item For RBBB, d1AVb, SB and ST, we \textit{accepted} all medical diagnoses. $26033$, $13645$, $12200$ and $14604$ diagnoses were \textit{accepted} in this fashion, respectively.
    \item For AF, we required not only that the exam was classified by the doctors as true, but also that the standard deviation of the NN intervals was higher than $646$. $14604$ diagnoses were \textit{accepted} using this rule.
\end{enumerate}
According to the sensitivity analysis,  the number of false positives that would be introduced  by this procedure was smaller than $3\%$ of the total number of exams.

\item After this process, we were still left with $34512$ exams where the corresponding diagnoses could neither be accepted nor rejected. These were \textit{manually reviewed} by medical students using the Telehealth ECG diagnostic system, under the supervision of a certified cardiologist with experience in ECG interpretation. The process of manually reviewing these ECGs took several months.
\end{enumerate}

It should be stressed that information from previous medical reports and automatic measurements were used only for obtaining the ground truth for training and validation sets and not on later stages of the DNN training. 

\subsection{Test set annotation}

The dataset used for testing the DNN was also obtained from TNMG's ECG system. It was independently annotated by two certified cardiologists with experience in electrocardiography. The kappa coefficients~\cite{cohen_coefficient_1960} indicate the inter-rater agreement for the two cardiologist and are: 0.741 for 1dAVb; 0.955 for RBBB; 0.964 for LBBB; 0.844 for SB; 0.831 for AF; 0.902 for ST. When they agreed, the common diagnosis was considered as ground truth. In cases where there was  \textit{any} disagreement, a third senior specialist, aware of the annotations from the other two, decided the diagnosis. The American Heart Association standardization~\cite{kligfield_recommendations_2007} was used as the guideline for the classification. 

It should be highlighted that the annotation was performed in an upgraded version of the TNMG software, in which the automatic measurements obtained by the Uni-G program are presented to the specialist, that has to choose the ECG diagnosis among a number of pre-specified classes of abnormalities. Thus, the diagnosis was codified directly into our classes and there was no need to extract the label from a textual report, as it was done for the training and validation sets.

\subsection{Neural network architecture and training}
\label{sec:Model}

\begin{figure}[ht]
    \centering
	\includegraphics[width=0.7\textwidth]{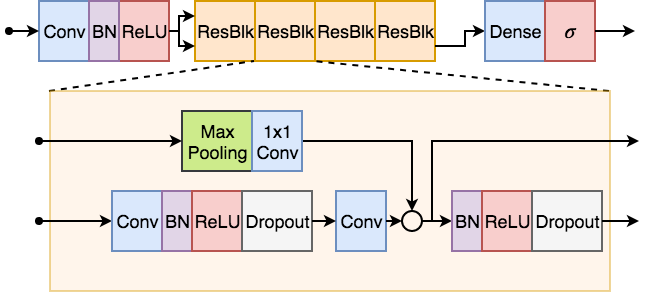}
    \caption{\textbf{(DNN architecture)} The uni-dimensional residual neural network architecture used for ECG classification.}
    \label{fig:Resnet}
\end{figure}

We used a convolutional neural network similar to the residual network~\cite{he_deep_2016}, but adapted to unidimensional signals. This architecture allows deep neural networks to be efficiently trained by including skip connections. We have adopted the modification in the  residual block proposed in~\cite{he_identity_2016}, which place the skip connection in the position displayed in Figure~\ref{fig:Resnet}. 

All ECG recordings are re-sampled to a 400 Hz sampling rate. The ECG recordings, which have between 7 and 10 seconds, are zero-padded resulting in a signal with 4096 samples for each lead. This signal is the input for the neural network.

The network consists of a convolutional layer (\texttt{Conv})  followed by $4$ residual blocks  with two convolutional layers per block.  The output of the last block is fed into a fully connected layer (\texttt{Dense}) with a sigmoid activation function, $\sigma$, which was used because the classes are not mutually exclusive (i.e. two or more classes may occur in the same exam). The output of each convolutional layer is rescaled using batch normalization, (\texttt{BN}),~\cite{ioffe_batch_2015} and fed into a rectified linear activation unit (\texttt{ReLU}). \texttt{Dropout}~\cite{srivastava_dropout_2014} is applied after the nonlinearity.

The convolutional layers have filter length 16, starting with 4096 samples and 64 filters for the first layer and residual block and increasing the number of filters by 64 every second residual block and subsampling by a factor of 4 every residual block.  \texttt{Max Pooling}~\cite{hutchison_evaluation_2010} and convolutional layers with filter length 1 (\texttt{1x1 Conv}) are included in the skip connections to make the dimensions match those from the signals in the main branch.

The average cross-entropy is minimized using the Adam optimizer~\cite{kingma_adam_2014} with default parameters and learning rate $\text{lr} = 0.001$. The learning rate is reduced by a factor of 10 whenever the validation loss does not present any improvement for 7 consecutive epochs. The neural network weights was initialized as in~\cite{he_delving_2015} and the bias were initialized with zeros. The training runs for 50 epochs with the final model being  the one with the best validation results during the optimization process.

\subsection{Hyperparameter tuning}

This final architecture and configuration of hyperparameters was obtained after approximately 30 iterations of the procedure: i) find the neural network weights in the training set; ii) check the performance in the validation set; and, iii) manually chose new hyperparameters and architecture using insight from previous iterations. We started this procedure from the set of hyperparameters and architecture used in~\cite{hannun_cardiologistlevel_2019}. It is also important to highlight that the choice of architecture and hyperparameters was done together with improvements in the dataset. Expert knowledge was used to take decision about how to incorporate, on the manual tuning procedure , information about previous iteration that were evaluated on slightly different versions of the dataset.

The hyperparameters were choosen among the following options: residual neural networks with $\{2, 4, 8, 16\}$ residual blocks, kernel size $\{8, 16, 32\}$, batch size $\{16, 32, 64\}$, initial learning rate $\{0.01, 0.001, 0.0001\}$, optimization algorithms \{SGD, ADAM\}, activation functions \{ReLU, ELU\}, dropout rate $\{0, 0.5, 0.8\}$, number of epochs without improvement in plateus between $5$ and $10$, that would result in a reduction in the learning rate between $0.1$ and $0.5$. Besides that, we also tried to: i) use vectorcardiogram linear transformation to reduce the dimensionality of the input; ii)  include LSTM layer before convolutional layers; iii) use residual network without the preactivation architecture proposed in~\cite{he_identity_2016}; iv) Use the convolutional architecture known as VGG; v) swiching the order of activation and batch normalization layer.

\subsection{Statistical and empirical analysis of test results}

We computed the precision-recall curve to assess the model discrimination of each rhythm class. This curve shows the relationship between precision (PPV) and recall (sensitivity), calculated using binary decision thresholds for each rhythm class. For imbalanced classes, such as our test set, this plot is more informative than the ROC plot~\cite{saito_precisionrecall_2015}. For the remaining analyses we fixed the DNN threshold to the value that maximized the F1 score, which is the harmonic mean between precision and recall. The F1 score was chosen here due to its robustness to class imbalance~\cite{saito_precisionrecall_2015}.

For the DNN with a fixed threshold, and for the medical residents and students, we computed the precision, the recall, the specificity, the F1 score and, also, the confusion matrix. This was done for each class. Bootstrapping~\cite{efron_introduction_1994} was used to analyze the empirical distribution of each of the scores: we generated 1000 different sets by sampling  with replacement from the test set, each set with the same number samples as in the test set, and computed the precision, the recall, the specificity and the F1 score for each. The resulting distributions are presented as a boxplot. We used the McNemar test~\cite{mcnemar_note_1947} to compare the misclassification distribution of the DNN and the medical residents and students on the test set and the kappa coefficient~\cite{cohen_coefficient_1960} to compare the inter-rater agreement.

All the misclassified exams were reviewed by an experienced cardiologist and, after an interview with the ECG reviewers, the errors were classified into: measurement errors, noise errors and unexplained errors (for the DNN only) and conceptual and attention errors (for medical residents and students only).

We evaluate the $F_1$ score for alternative scenarios where we use 90\%-5\%-5\% splits of the 2,322,513 records. With the splits ordered: randomly; by date; and, stratified by patients. The neural networks developed in these alternative scenarios are evaluated on both the original test set (n=827) and on the additional test splits (last 5\% split).  The distribution of the performance in each scenario is computed by a bootstrap analysis (with 1000 and 200 samples, respectively) and the resulting boxplots are displayed in the supplementary material.

\subsection*{Data availability} The test dataset used in this study is openly available, and can be downloaded at\\ 
\verb|[https://doi.org/10.5281/zenodo.3625006]| The weights of all deep neural network models we developed for this paper are  available at \verb|[https://doi.org/10.5281/zenodo.3625017]|. Restrictions apply to the availability of the training set. Requests to access the training data will be considered on an individual basis by the Telehealth Network of Minas Gerais. Any data use will be restricted to non-commercial research purposes, and the data will only be made available on execution of appropriate data use agreements. The source data underlying Supplementary Figures 1 and 2 are provided as a Source Data file.

\subsection*{Code availability} The code for  training and evaluating the DNN model, and, also, for generating figures and tables in this paper, is available at: \verb|[https://github.com/antonior92/automatic-ecg-diagnosis]|. 

\subsection*{Research ethics statement} This study complies with all relevant ethical regulations. It was approved by the Research Ethics Committee of the Universidade Federal de Minas Gerais, protocol 68496317.7.0000.5149.

\subsection*{Acknowledgments}
This research was partly supported by the Brazilian Agencies CNPq, CAPES, and FAPEMIG, by projects IATS, MASWeb, INCT-Cyber and Atmosphere, and by the \emph{Wallenberg AI, Autonomous Systems and Software Program (WASP)} funded by Knut and Alice Wallenberg Foundation. We also thank NVIDIA for awarding our project with a Titan V GPU. ALR and WMJr are recipients of unrestricted research scholarships from CNPq; AHR receives a scholarship from CAPES and CNPq; and, MHR and DMO receives a Google Latin America Research Award scholarship. None of the funding agencies had any role in the design, analysis or interpretation of the study.

\subsection*{Author contribution statement}
A.H.R., M.H.R., G.P., D.M.O., P.R.G., J.A.C, M.P.S.F and A.L.R were responsible for the study design. A.L.R conceived the project and acted as project leader. A.H.R., M.H.R and C.A.  choose the architecture, implemented and tuned the deep neural network. A.H.R did the statistical analysis of the test data and generated the figures and tables. M.H.R., G.M.M.P, J.A.C. were responsible for the preprocessing and annotating the datasets. G.M.M.P was responsible for the error analysis. D.M.O. implemented the semi-supervised methodology to extract the text label. P.R.G. implemented the user interface used to generate the dataset. P.R.G. and M.P.S.F were responsible for maintenance and extraction of the database.  P.W.M., W.M.Jr., and T.B.S helped in the interpretation of the data. A.H.R., M.H.R, P.W.M., T.B.S. and A.L.R. contributed to the writing and all authors revised it critically for important intellectual content. All authors read and approved the submitted manuscript.

\subsection*{Competing interesting statement}
The authors declare no competing interests.


\newpage

{\raggedright\baselineskip= 24pt\titlefont Supplementary Information \par}
\vskip40pt

\renewcommand{\figurename}{Supplementary Figure}
\renewcommand{\thefigure}{\arabic{figure}}
\setcounter{figure}{0}

\renewcommand{\tablename}{Supplementary Table}
\renewcommand{\thetable}{\arabic{table}}
\setcounter{table}{0}

\begin{table}[h]
    \centering
    \pgfkeys{/pgf/number format/.cd,fixed,precision=0, fixed zerofill=true}
    \makebox[\textwidth][c]{ 
    \begin{filecontents*}{tables/confusion_matrices.csv}
predictor,,DNN,DNN,cardio.,cardio.,emerg.,emerg.,stud.,stud.
predicted label,,not present,present,not present,present,not present,present,not present,present
1dAVb,not present,795,4,797,2,786,13,782,17
,present,2,26,9,19,5,23,2,26
RBBB,not present,789,4,788,5,792,1,790,3
,present,0,34,1,33,8,26,2,32
LBBB,not present,797,0,797,0,796,1,795,2
,present,0,30,3,27,4,26,3,27
SB,not present,808,3,808,3,808,3,807,4
,present,1,15,1,15,2,14,4,12
AF,not present,814,0,811,3,812,2,805,9
,present,3,10,3,10,5,8,1,12
ST,not present,788,2,789,1,788,2,787,3
,present,1,36,7,30,2,35,6,31
\end{filecontents*}
\pgfplotstabletypeset[ 
      skip first n=1,
      every nth row={2[-1]}{before row={\rowcolor[gray]{0.9}}, after row=\cline{2-10}},
      display columns/0/.style={column name = , string type, column type/.add={}{}},
      display columns/1/.style={column name = \cellcolor[gray]{0.8} true label, string type},
      display columns/2/.style={column name = not present, column type/.add={|}{}},
      display columns/3/.style={column name = present},
      display columns/4/.style={column name = not present, column type/.add={|}{}},
      display columns/5/.style={column name = present},
      display columns/6/.style={column name = not present, column type/.add={|}{}},
      display columns/7/.style={column name = present},
      display columns/8/.style={column name = not present, column type/.add={|}{}},
      display columns/9/.style={column name = present},
      col sep=comma, 
      every last row/.style={ after row=\bottomrule},
      every head row/.style={before row={
      \multicolumn{1}{c}{}& \multicolumn{1}{c}{}& \multicolumn{8}{c}{\cellcolor[gray]{0.8}predicted label}\\
      \multicolumn{2}{c}{}& \multicolumn{2}{c}{DNN} & \multicolumn{2}{c}{cardio.} & \multicolumn{2}{c}{emerg.} & \multicolumn{2}{c}{stud.}\\}, after row=\midrule}
    ] {tables/confusion_matrices.csv}}
    \caption{\textbf{(Confusion matrices)} Show the absolute number of: i) false posives; ii) false negatives; iii) true positives; and, iv) true negatives, for each abnormality on the test set.}
    \label{tab:confusion_matrices}
\end{table}

\begin{table}[h]
    \centering
    \pgfkeys{/pgf/number format/.cd,fixed,precision=3, fixed zerofill=true}
    \subfloat[][]{
    \begin{filecontents*}{tables/kappa.csv}
,BAV1,BRD,BRE,BradSin,FibAtrial,TaquiSin
DNN vs cardio.,0.656,0.917,0.945,0.830,0.780,0.864
DNN vs emerg.,0.684,0.792,0.909,0.796,0.595,0.930
DNN vs stud.,0.642,0.928,0.912,0.760,0.574,0.855
cardio. vs emerg.,0.656,0.824,0.923,0.912,0.515,0.847
cardio. vs stud.,0.612,0.871,0.889,0.880,0.700,0.792
emerg. vs stud.,0.615,0.799,0.852,0.907,0.508,0.897
\end{filecontents*}
\pgfplotstabletypeset[ 
      col sep=comma, 
      display columns/0/.style={column name = , string type},
      display columns/1/.style={column name = 1dAVb},
      display columns/2/.style={column name = RBBB},
      display columns/3/.style={column name = LBBB},
      display columns/4/.style={column name = SB},
      display columns/5/.style={column name = AF},
      display columns/6/.style={column name = ST},
      every last row/.style={ after row=\bottomrule},
      every head row/.style={before row=\toprule, after row=\midrule}
    ]{tables/kappa.csv}}\\
        \subfloat[][]{
    \begin{filecontents*}{tables/kappas_annotators_and_DNN.csv}
,BAV1,BRD,BRE,BradSin,FibAtrial,TaquiSin
DNN vs Cert. cardiol. 1,0.758,0.928,0.964,0.770,0.696,0.847
DNN vs Certif. cardiol. 2,0.852,0.942,1.000,0.770,0.746,0.884
Cert. cardiol. 1 vs Certif. cardiol. 2,0.741,0.955,0.964,0.844,0.831,0.902
\end{filecontents*}
\pgfplotstabletypeset[ 
      col sep=comma, 
      display columns/0/.style={column name = , string type},
      display columns/1/.style={column name = 1dAVb},
      display columns/2/.style={column name = RBBB},
      display columns/3/.style={column name = LBBB},
      display columns/4/.style={column name = SB},
      display columns/5/.style={column name = AF},
      display columns/6/.style={column name = ST},
      every last row/.style={ after row=\bottomrule},
      every head row/.style={before row=\toprule, after row=\midrule}
    ]{tables/kappas_annotators_and_DNN.csv}}
    \caption{\textbf{(Kappa coefficients)} Show the Kappa scores measuring the inter-rater agreement on the test set. In (a), we compare the DNN, the medical residents and the students two at a time. In (b), we compare the DNN, and the certified cardiologists that annotated the test set (\texttt{certif. cardiol.}). If the raters are in complete agreement then it is equal to 1. If there is no agreement among the raters other than what would be expected by chance it is equal to 0.}
    \label{tab:kappa}
\end{table}

\begin{figure}[h]
    \centering
    \subfloat[][Precision (PPV)]{\includegraphics[width=0.3\textwidth]{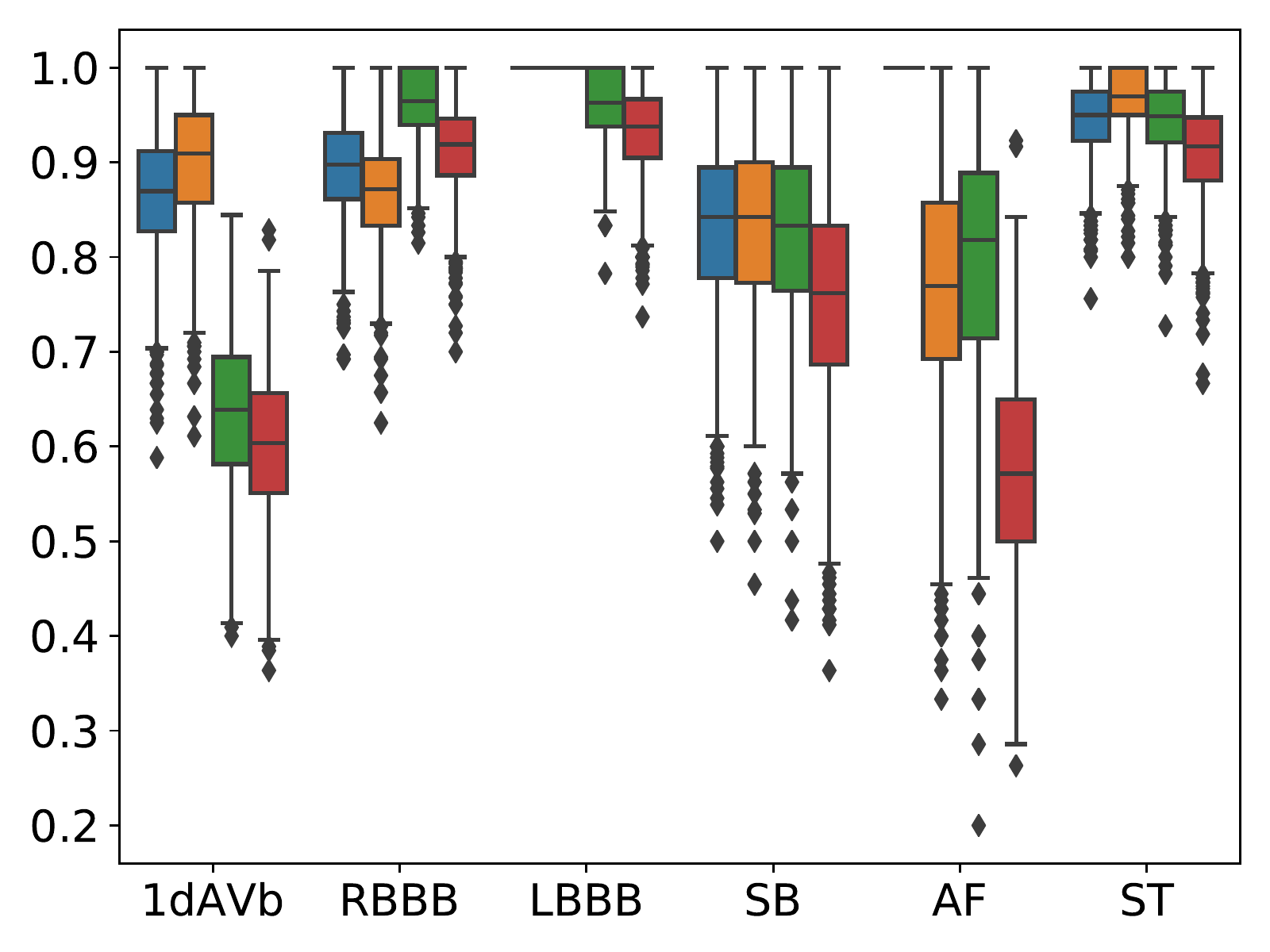}}
    \subfloat[][Recall (Sensitivity)]{\includegraphics[width=0.3\textwidth]{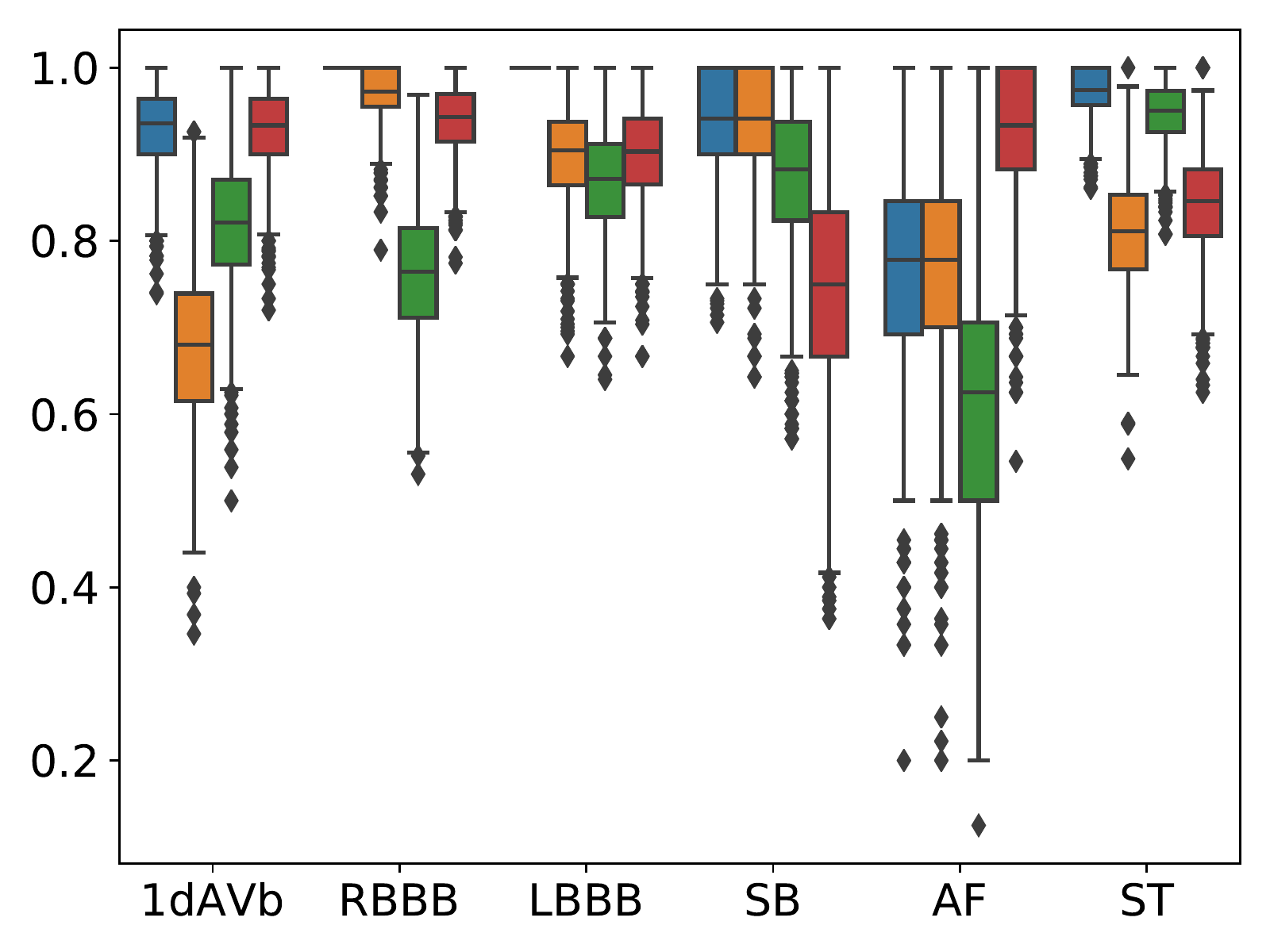}}
    \subfloat[][Specificity]{\includegraphics[width=0.3\textwidth]{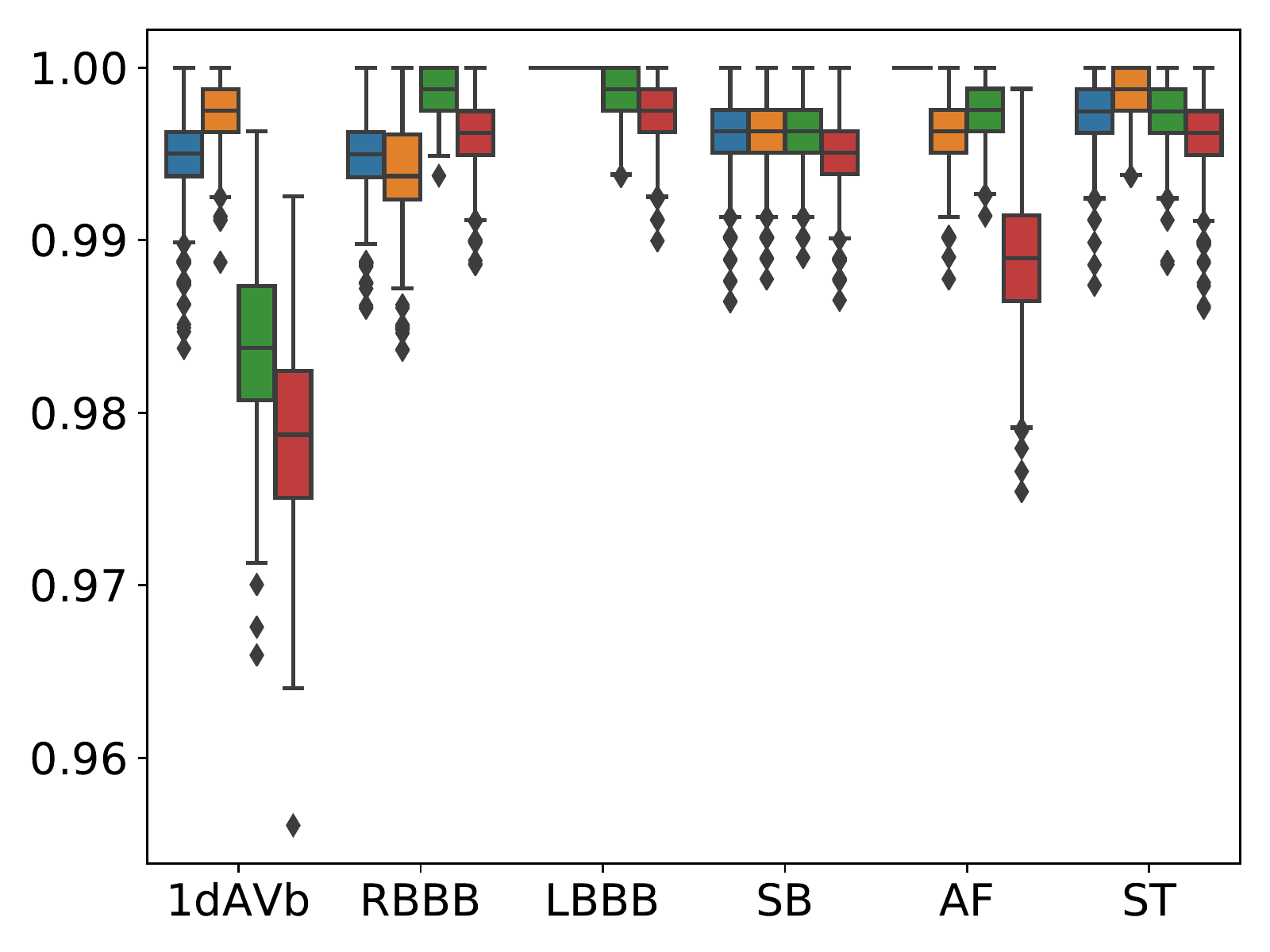}}\\
    \subfloat[][F1 score]{\includegraphics[width=0.45\textwidth]{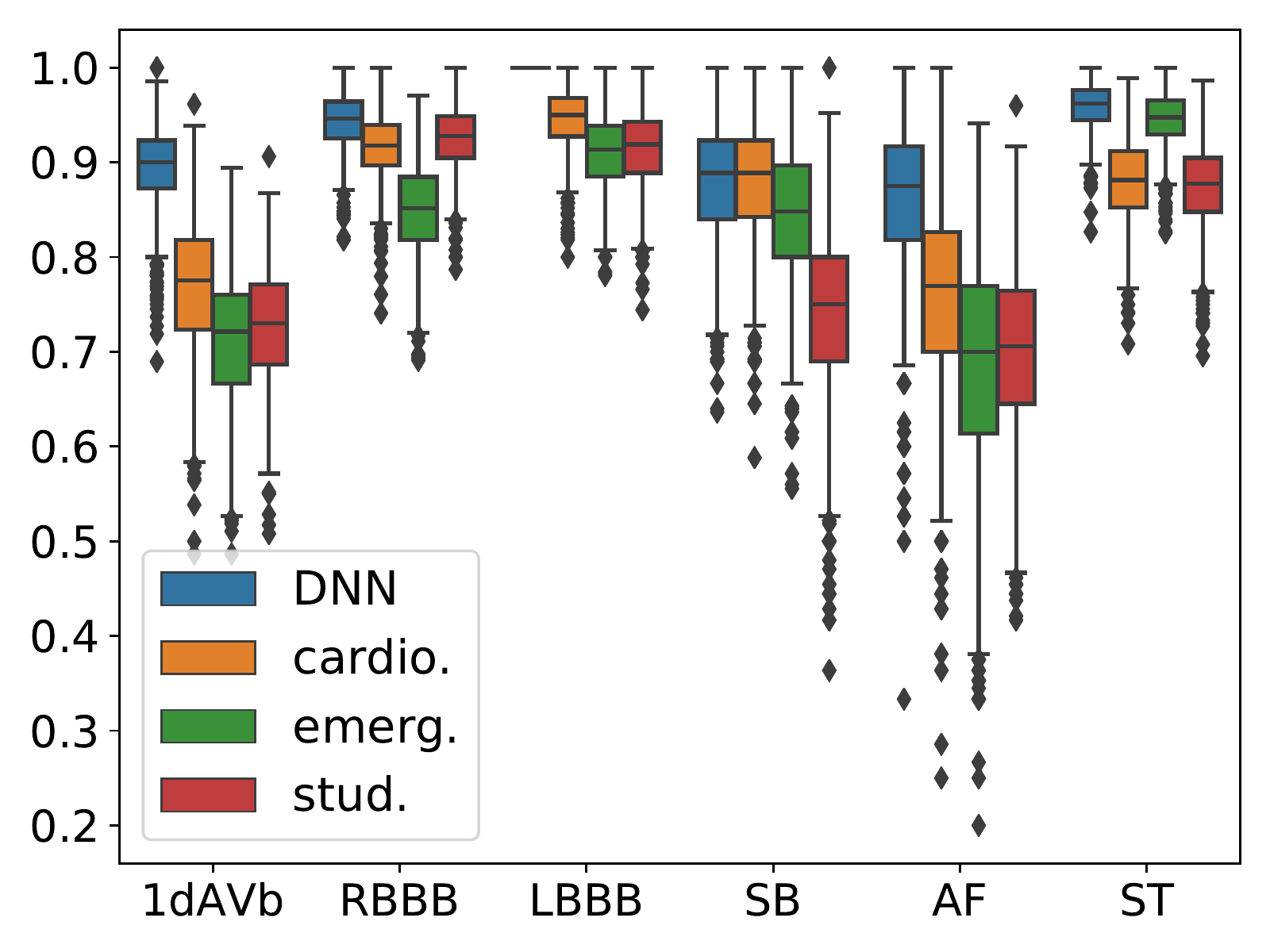}}
    \caption{\textbf{(Bootstrapped scores)} Display boxplots of empirical distribution of precision, recall, specificity and F1 score on the test set. Sampling with replacement (i.e. bootstrapping) from the test set was used to generate $n=1000$ samples. The results are given for the DNN, the medical residents and students. Source data are provided as a Source Data file. The boxplots should be read as follows: the central line correspond to the median value of the empirical distribution, the box region correspond to the range of values between the first and third quartile (also knonw as interquartile range or IQR), the whiskers extend from 1.5 IQR below and above the firsth and third quartiles, values outside of that range are considered outliers and show as diamonds. }
    \label{fig:bootstrap}
\end{figure}

\begin{table}[h]
    \centering
    \pgfkeys{/pgf/number format/.cd,fixed,precision=3, fixed zerofill=true}
    \makebox[\textwidth][c]{ 
    \begin{filecontents*}{tables/mcnemar.csv}
,BAV1,BRD,BRE,BradSin,FibAtrial,TaquiSin
DNN vs cardio.,0.225,0.414,0.083,1.000,0.180,0.096
DNN vs emerg.,0.007,0.166,0.025,0.705,0.157,0.655
DNN vs stud.,0.009,0.655,0.025,0.157,0.052,0.058
cardio. vs emerg.,0.108,0.366,0.317,0.564,0.763,0.206
cardio. vs stud.,0.102,0.739,0.414,0.046,0.206,0.782
emerg. vs stud.,0.853,0.248,1.000,0.083,0.439,0.059
\end{filecontents*}
\pgfplotstabletypeset[ 
      col sep=comma, 
      display columns/0/.style={column name = , string type},
      display columns/1/.style={column name = 1dAVb},
      display columns/2/.style={column name = RBBB},
      display columns/3/.style={column name = LBBB},
      display columns/4/.style={column name = SB},
      display columns/5/.style={column name = AF},
      display columns/6/.style={column name = ST},
      every last row/.style={ after row=\bottomrule},
      every row 1 column 1/.style={postproc cell content/.style={@cell content=\textbf{##1}}},
      every row 2 column 1/.style={postproc cell content/.style={@cell content=\textbf{##1}}},
      every row 4 column 4/.style={postproc cell content/.style={@cell content=\textbf{##1}}},
      every head row/.style={before row=\toprule, after row=\midrule}
    ]{tables/mcnemar.csv}
    }
    \caption{\textbf{(McNemar test)} Display the \textit{p}-values for the McNemar test comparing the misclassification on the test set. The DNN, the medical residents and the students were compared two at a time. Entries with statistical significance (with 0.05 significance level) are displayed in \textbf{boldface}. }
    \label{tab:mcnemar}
\end{table}

\begin{figure}[h]
    \centering
    \subfloat[][original test set]{\includegraphics[width=0.5\textwidth]{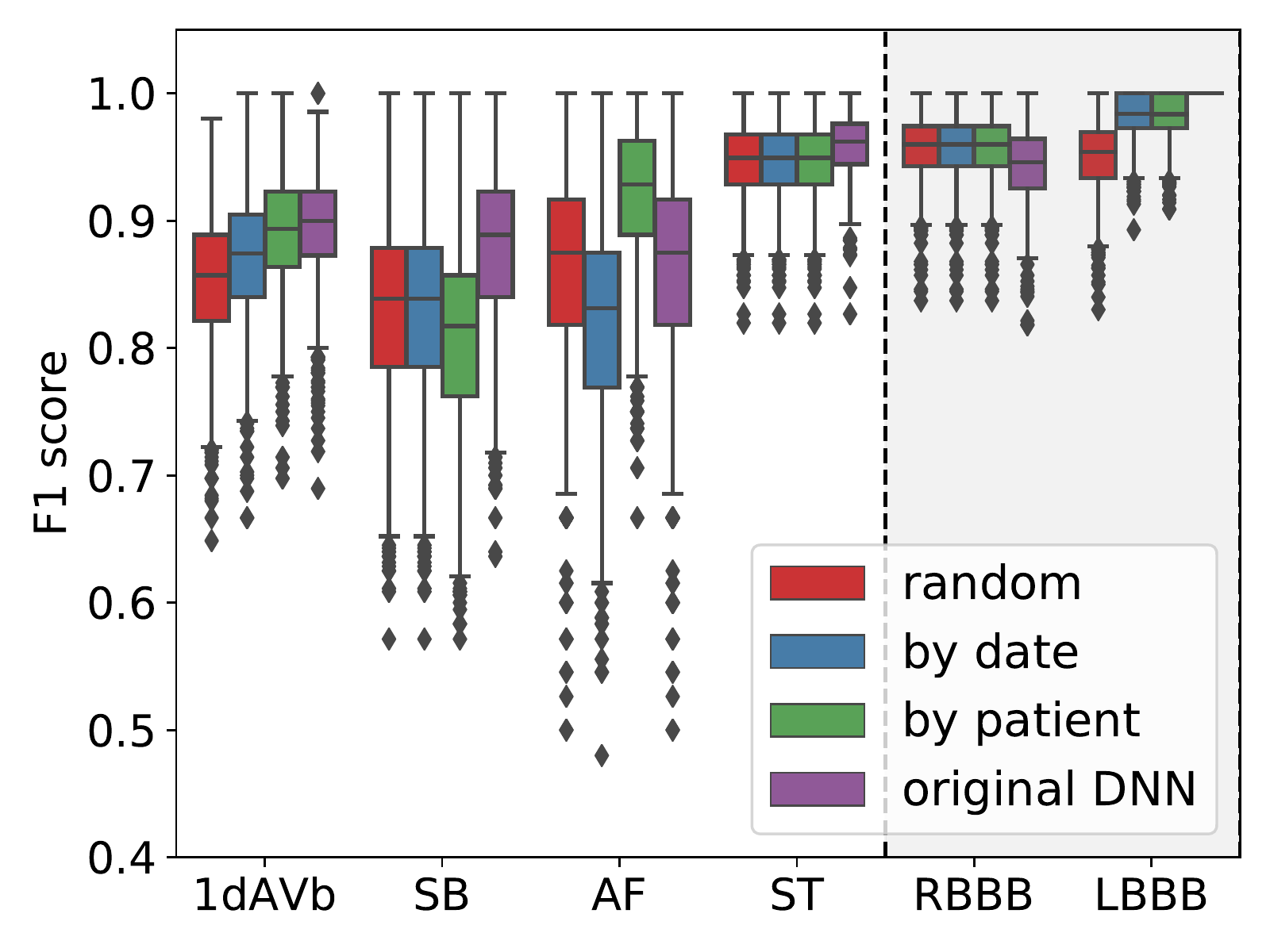}}
    \subfloat[][secondary test set (last $5\%$ split)]{\includegraphics[width=0.5\textwidth]{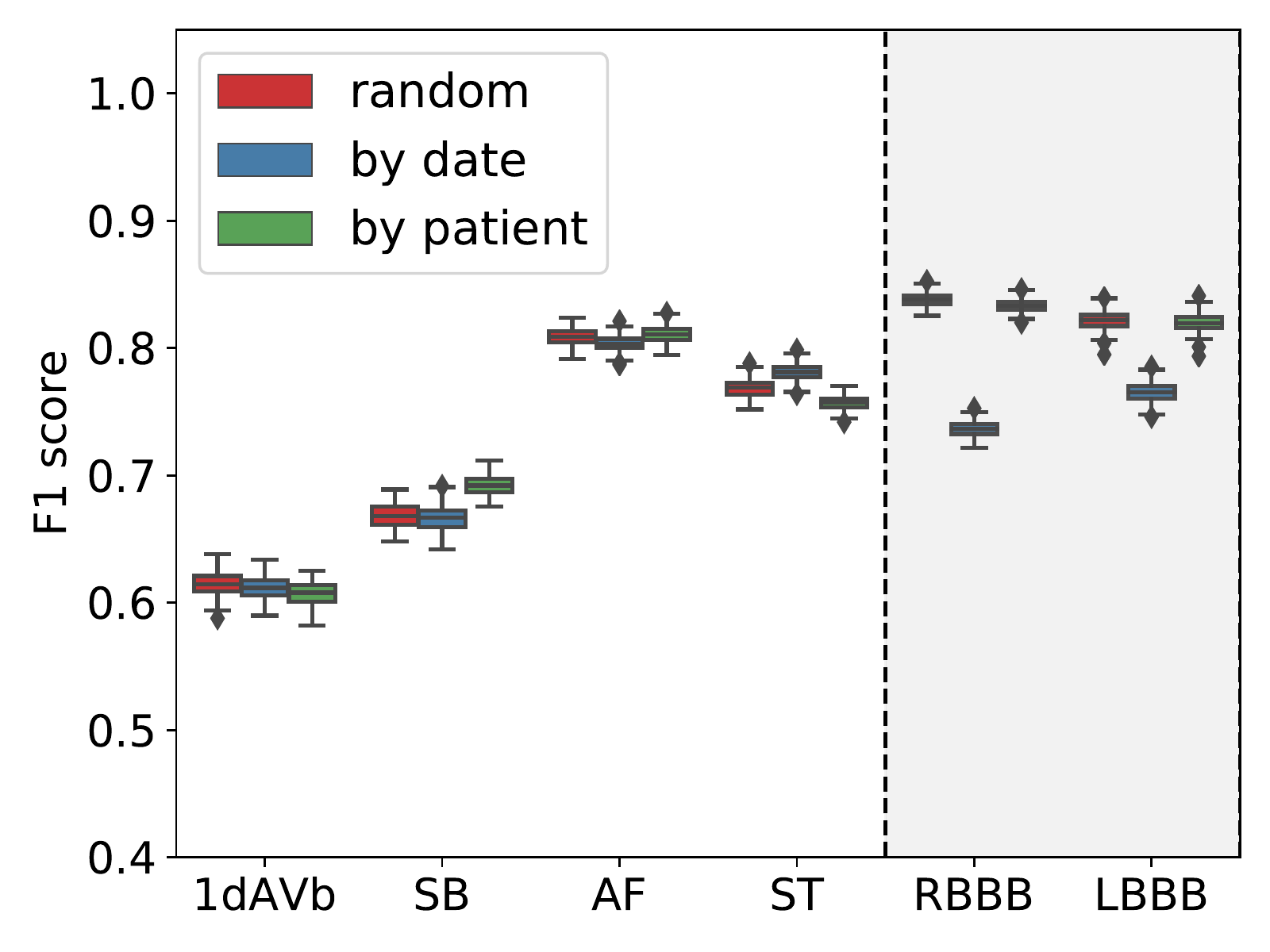}}
    \caption{\footnotesize{\textbf{(Bootstrapped scores for alternative splits)} Boxplots for the bootstrapped $F_1$ scores for the DNN using alternative 90\%-5\%-5\% splits for training, validation and as a \textit{secondary} test set. For the splits ordered: randomly; by date; and, stratified by patients. In all cases, the performance is evaluated on: (a) the original test set for $n=1000$ bootstraped samples; and, on (b) the secondary test set (last $5\%$ split) for $n=200$ bootstraped samples.   On (a), we also present the original DNN performance for comparison, which was developed using a 98\%-2\% split. The performance gap between (a) and (b) is due to the difference in the gold standard. The secondary test set obtained from the last $5\%$ has a less accurate annotation, since it has not been annotated by multiple doctors and it uses natural language processing to extract the diagnosis from a written report. This extra noise result in worse $F_1$ score in (b) when compared with (a). On the other hand, the secondary 5\% test split contain more than 100,000 records, which yield more stable performance in the bootstrap analysis, with more concentrated empirical distributions for the $F_1$ score. Both RBBB and LBBB (highlighted on the plot) present on (b) a statistically significant difference between the performance of the split ordered by date and the other two splits, that difference is due to some changes in personal that took place in the the Telehealth center, that affected the period used in the test set (10-2016 to 06-2017), resulting in lower annotation quality. A certified cardiologist reviewed  cases for which the neural network have been considered wrong when compared to the gold standard from the 5\% split collected from 10-2016 to 06-2017, 100 supposedly wrong RBBB and 100 supposedly wrong LBBB. The certified cardiologist reported that the neural network is actually correct, respectively, $86\%$ and $83\%$ percent of the cases. This analysis show the importance of a test set with a good annotation quality to obtain reliable estimation of the DNN performance. And, also, that periods of lower annotation quality in the dataset are overcome by a very high number of examples. Source data are provided as a Source Data file. See Supplementary Figure~\ref{fig:bootstrap} caption for the definition of all elements in the boxplot.}}
    \label{fig:boxplot_F1_score_on_newtest}
\end{figure}

\begin{figure}[h]
    \centering
    \subfloat[][]{\includegraphics[width=0.5\textwidth]{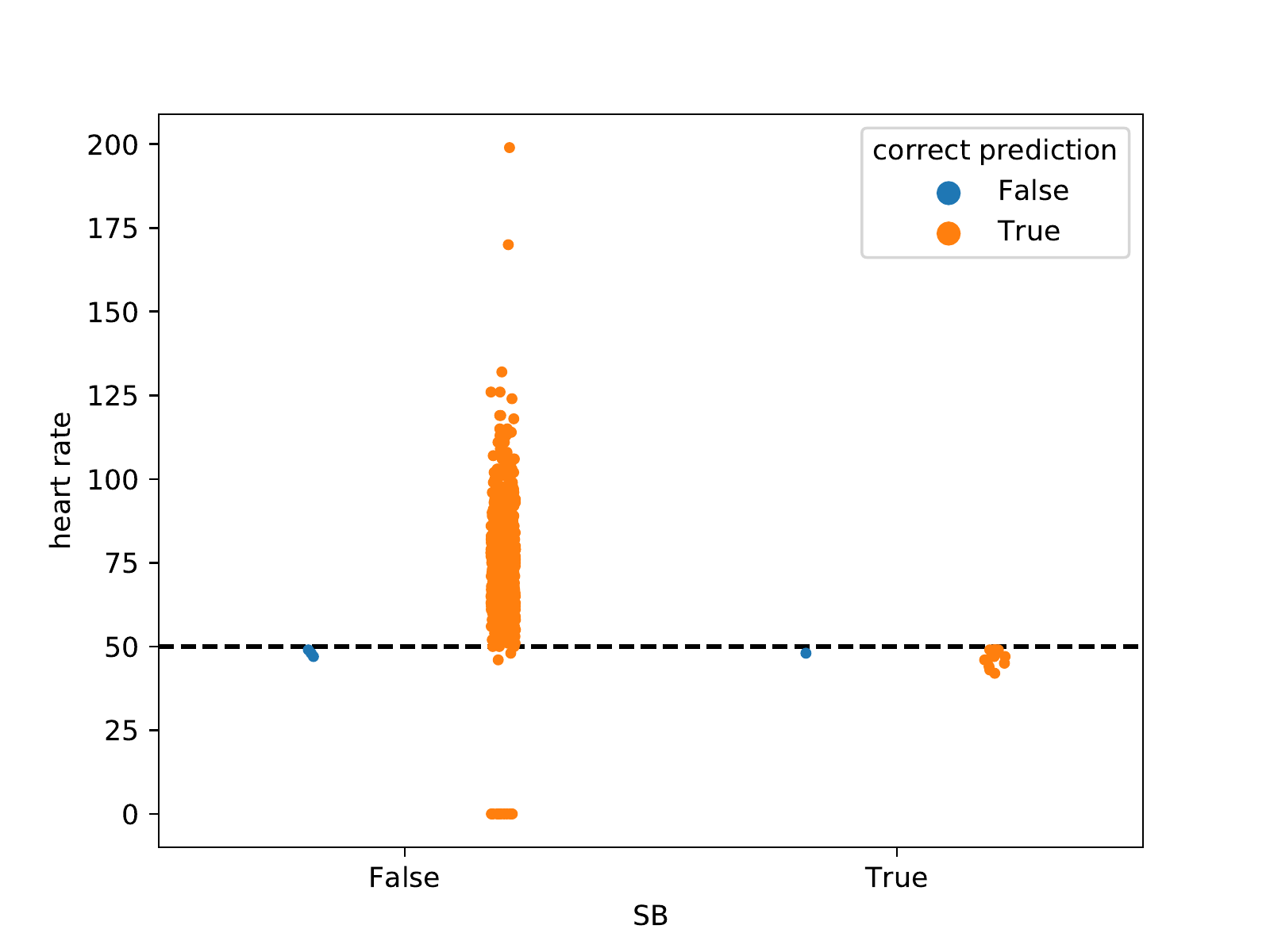}}
    \subfloat[][]{\includegraphics[width=0.5\textwidth]{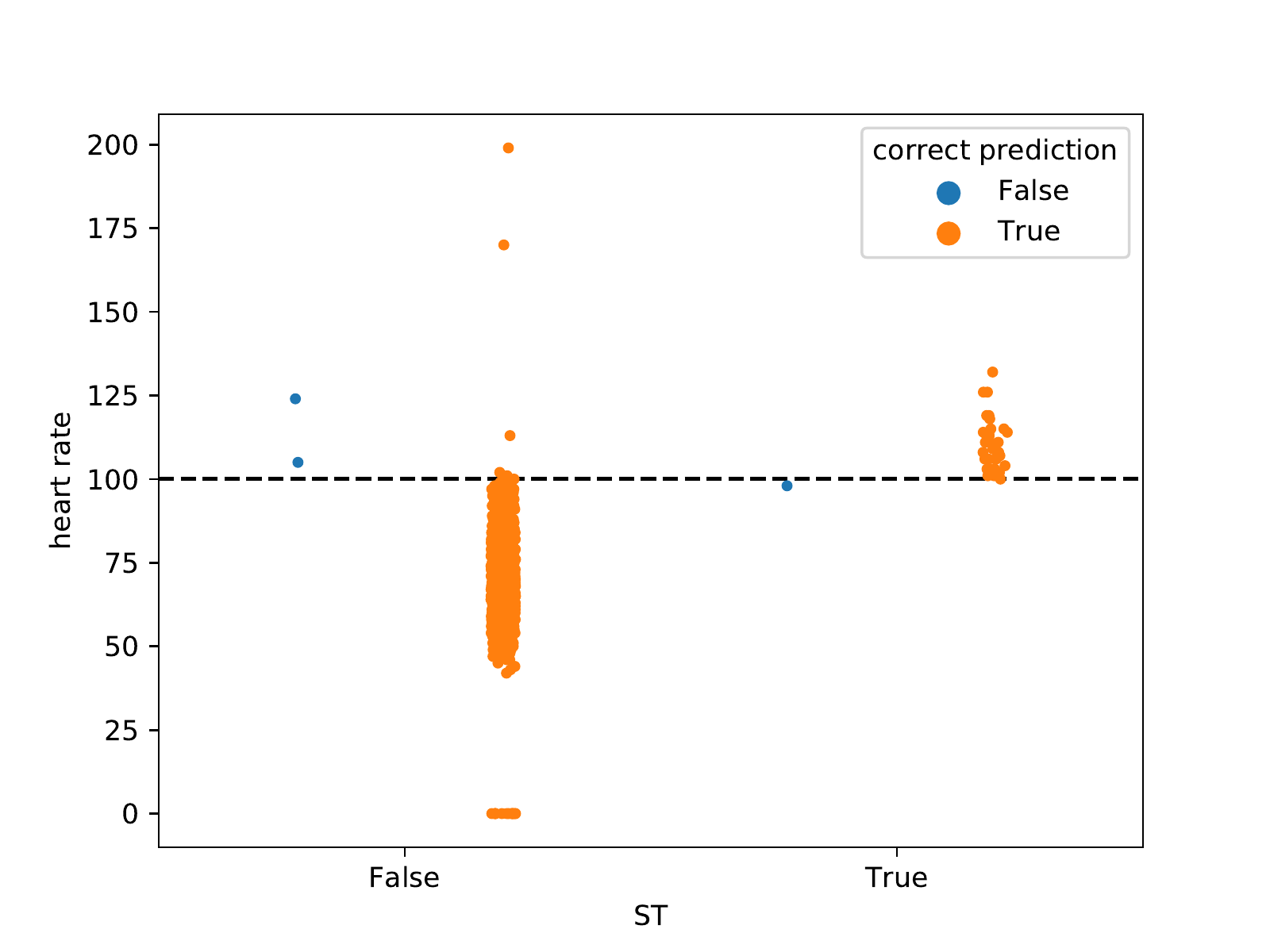}}
    \caption{\textbf{(Heart rate \textit{vs} DNN predictions)} Heart rate measured by the Uni-G software for samples in the test set is given on the $y$-axis. The color indicates if the \textit{DNN} make the correct prediction or not. The x-axis separates the dataset accordingly to the presence of: SB in (a); and, ST in (b). A horizontal line show the threshold of 50 bpm for SB in (a); and, of 100 bpm for ST in (b), which delimit the consensus definition of SB and ST. Notice that most exams for which the neural network fails to get the correct result are very close to this threshold line and are the borderline cases we mentioned in the discussion. It should be highlighted that this automatic measurement system is not perfect, and measurements that may indicate some of the conditions  do not necessarily agree with our board of cardiologist (e.g. there are exams with heart rate above 100 acording to Uni-G that are not classified by our cardiologist as ST).}
    \label{fig:heart_rate_vs_mistakes}
\end{figure}
\end{document}